\documentclass[sigconf]{acmart}

\usepackage{multirow} 
\usepackage{graphicx}
\usepackage{ragged2e}
\usepackage{array}
\usepackage{longtable}
\usepackage{rotating}
\usepackage{balance}

\usepackage[justification=centering]{caption}

\AtBeginDocument{%
  }

\setcopyright{acmlicensed}
\copyrightyear{2018}
\acmYear{2018}
\acmDOI{XXXXXXX.XXXXXXX}

\acmConference[Conference acronym 'XX]{Make sure to enter the correct
  conference title from your rights confirmation emai}{June 03--05,
  2018}{Woodstock, NY}
\acmISBN{978-1-4503-XXXX-X/18/06}

\begin{document}

\title{GinAR: An End-To-End Multivariate Time Series Forecasting Model Suitable for Variable Missing}


\author{Chengqing Yu}
\author{Fei Wang}
\authornote{Fei Wang and Yongjun Xu are the corresponding authors.}
\author{Zezhi Shao}
\affiliation{%
  \institution{Institute of Computing Technology, }
  \city{Chinese Academy of Sciences, Beijing}
  \country{China}
}
\affiliation{%
  \institution{University of Chinese Academy of Sciences,}
  \city{Beijing}
  \country{China}
}
\email{{yuchengqing22b, wangfei, Shaozezhi19b}@ict.ac.cn}

\author{Tangwen Qian}
\author{Zhao Zhang}
\affiliation{%
  \institution{Institute of Computing Technology,}
  \city{Chinese Academy of Sciences, Beijing}
  \country{China}
}
\email{{qiantangwen, zhangzhao2021}@ict.ac.cn}

\author{Wei Wei}
\affiliation{%
  \institution{School of Computer Science and Technology, Huazhong University of Science and Technology,}
  \city{Wuhan}
  \country{China}
}
\email{weiw@hust.edu.cn}

\author{Yongjun Xu}
\authornotemark[1]
\affiliation{%
  \institution{Institute of Computing Technology, }
  \city{Chinese Academy of Sciences, Beijing}
  \country{China}
}
\email{xyj@ict.ac.cn}

\renewcommand{\shortauthors}{Yu et al.}

\begin{abstract}
Multivariate time series forecasting (MTSF) is crucial for decision-making to precisely forecast the future values/trends, based on the  complex relationships identified from historical observations of multiple sequences. Recently, Spatial-Temporal Graph Neural Networks (STGNNs) have gradually become the theme of MTSF model as their powerful capability in mining spatial-temporal dependencies, but almost of them heavily rely on the assumption of historical data integrity. In reality, due to factors such as data collector failures and time-consuming repairment, it is extremely challenging to collect the whole historical observations without missing any variable. In this case, STGNNs can only utilize a subset of normal variables and easily suffer from the incorrect spatial-temporal dependency modeling issue, resulting in the degradation of their forecasting performance. To address the problem, in this paper, we propose a novel Graph Interpolation Attention Recursive Network (named GinAR) to precisely model the spatial-temporal dependencies over the limited collected data for forecasting. In GinAR, it consists of two key components, that is, interpolation attention and adaptive graph convolution to take place of the fully connected layer of  simple recursive units, and thus are capable of recovering all missing variables and reconstructing the correct spatial-temporal dependencies for recursively modeling of multivariate time series data, respectively. Extensive experiments conducted on five real-world datasets demonstrate that GinAR outperforms 11 SOTA baselines, and even when 90\% of variables are missing, it can still accurately predict the future values of all variables.
\end{abstract}




\maketitle

\section{Introduction}
Multivariate time series forecasting (MTSF) is widely used in practice, such as transportation ~\cite{shang2022new}, environment ~\cite{tan2022new} and others ~\cite{xu2021artificial}. It predicts future values of multiple interlinked time series by using their historical observations, and contributes to decision-making ~\cite{wang2023ai, xu2023artificial}. Indeed, multivariate time series (MTS) can be formalized as a kind of classical spatial-temporal graph data ~\cite{cao2020spectral}, such as traffic flow ~\cite{cirstea2022towards}, each variable of which is collected in chronological order, using a sensor deployed at an independent position. Naturally, they usually have two key factors, temporal dependency ~\cite{wang2023learning} and spatial correlation ~\cite{tang2023recurrent}. The former characterizes complex patterns (e.g., causal relationships) of instances in chronological order, and the later depicts the differences of time series corrected each other in spatial dimension. Therefore, effectively mining spatial-temporal dependencies is crucial for MTSF to precisely predict future values of the time series, or to better understand of how they interact \cite{qian2023adaptraj, sun2022human}.

\begin{figure}
  \centering
  \includegraphics[width=\linewidth]{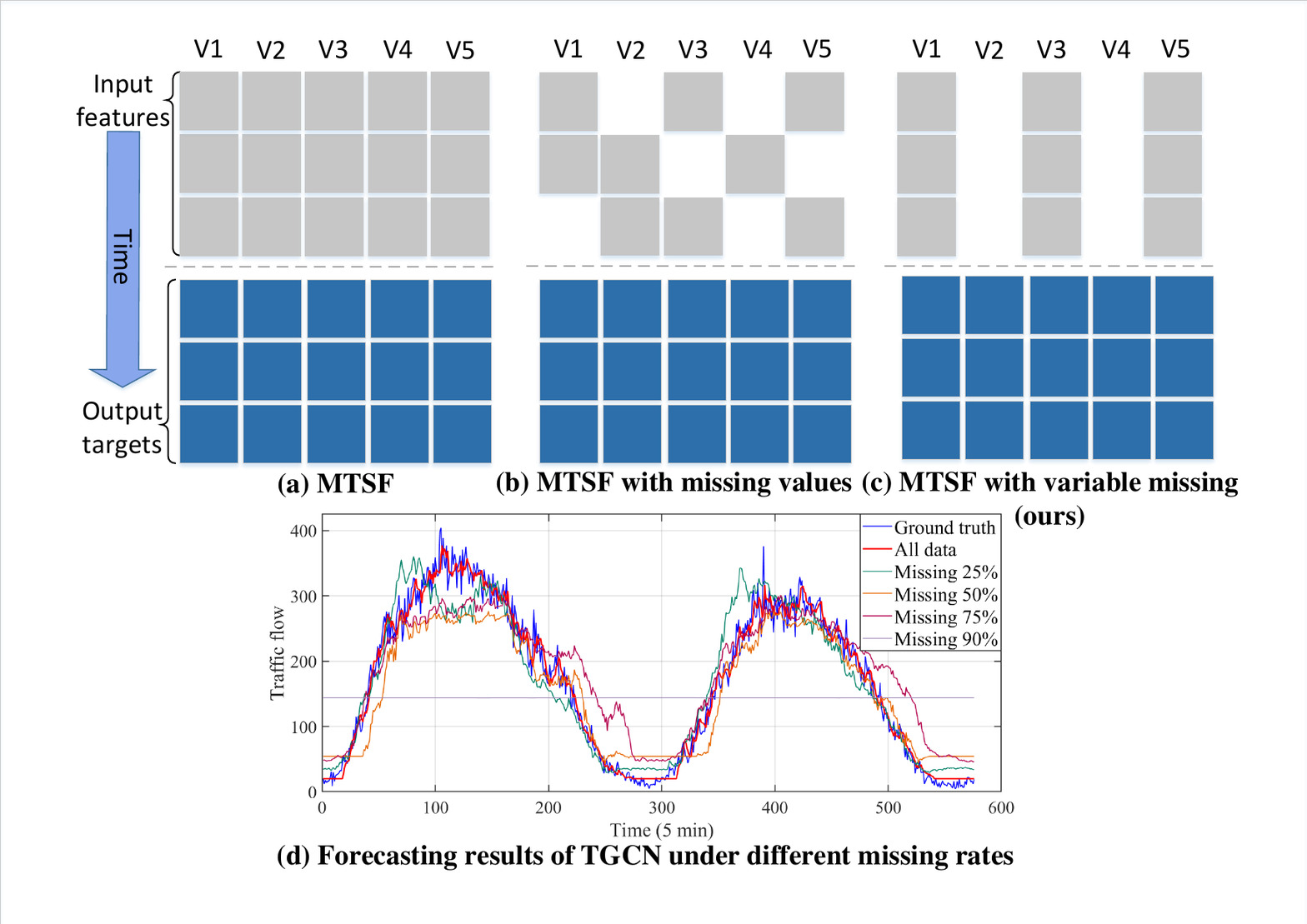}
  \caption{The principle and examples of multivariate time series forecasting with variable missing. V1 to V5 represent different variables. Compared to the other two tasks, our task can only use historical observations of certain variables to predict the future values of all variables. The forecasting performance of TGCN declines as the missing rate increases.}
  \label{fig1}
\end{figure}

Recently, Spatial-Temporal Graph Neural Networks (STGNNs) combine the sequence model and graph convolution (GCN) to capture spatial-temporal dependencies of MTS and achieve significant progress in MTSF ~\cite{chen2023higher}, but their superior performances heavily rely on the data quantity ~\cite{zhou2023sloth}. Since the time series data in practice is always incomplete, it is very challenging to obtain whole historical observations of all variables for accurate forecasting ~\cite{luo2019e2gan}. To make things worse, the data from some variables may be even unavailable for a long time under certain conditions ~\cite{chen2022novel}. We can take a classical MTS application (i.e., air quality forecasting) as an example, the data collectors may easily work anomaly owing to some unforeseen factors (e.g., horrible weather) ~\cite{pachal2022sequence}. Because equipment maintenance usually takes days or even months, corresponding data collectors only output outliers for a long time ~\cite{yick2008wireless}. Thus, STGNNs need to address a problem, namely, whole history observations missing of some variables. This means that STGNNs only achieve MTSF using the remaining normal variables (shown in \autoref{fig1} (c)), which severely limits their performance. To alleviate this problem, some works ~\cite{chauhan2022multi} only predict values of remaining normal variables by discarding all missing variables. However, if missing variables are key samples (e.g., important locations like hub nodes), the inability to predict their values will profoundly affect decision-making ~\cite{wei2023lstm}. 

The above phenomenon shows that MTSF faces a significant practical challenge: \textbf{how to forecast MTS when missing part of variables?} By rethinking the characteristics of STGNNs and this task, the main problem is that STGNNs easily capture incorrect spatial-temporal dependencies during the modeling process, resulting in error accumulation and degraded forecasting performance. 
On the one hand, since each missing variable is usually a straight-line sequence composed of outliers, the sequence model in STGNNs ~\cite{chowdhury2022tarnet} cannot mine any valuable pattern and information, resulting in incorrect temporal dependencies. On the other hand, existing STGNNs ~\cite{hu2022mgcn,deng2024disentangling,li2023dynamic} need to use historical observations from all variables to construct spatial correlations. Because whole history observations of some variables are missing, existing STGNNs cannot establish spatial correlations between missing and normal variables, leading to incorrect spatial correlations. In this case, as the missing rate increases, the above phenomena become more serious, leading to a significant decline in the performance of STGNNs. For example, a classic STGNN model, temporal graph convolutional network (TGCN) ~\cite{zhao2019t}, is used for further analysis when given different missing rates on PEMS04. \autoref{fig1} (d) shows that its performance deteriorates while increasing the missing rate.

At present, an intuitive strategy for addressing this challenge is to combine imputation and forecasting methods and propose two-stage models ~\cite{li2023multivariate}. However, classic imputation methods ~\cite{RN18,chen2021low} primarily rely on the context information of time series to recover missing values. When history observations from some variables are unavailable for a long time, these methods cannot achieve reliable recovery effects since the missing variables do not have any normal value in the temporal dimension. In addition to the above classical methods, existing mainstream imputation methods ~\cite{shan2023nrtsi, blazquez2023selective} combine the context information and spatial correlations of MTS for generating plausible missing values. However, they also have two problems: (1) Components that use context information in these imputation methods also introduce incorrect temporal dependencies, limiting the effectiveness of data recovery and leading to error accumulation ~\cite{chauhan2022multi}. (2) these imputation methods mainly rely on fixed spatial correlations (such as road network structure) to establish correspondences between missing variables and normal variables ~\cite{ye2021spatial}. When the missing rate is significant, they cannot fully use all normal variables to recover missing variables, resulting in the ineffective recovery of missing variables that do not correspond with normal variables ~\cite{yu2023MRIformer}. In general, due to the introduction of incorrect temporal dependencies and the lack of sufficient correspondences between missing variables and normal variables, two-stage models cannot work well in MTSF with variable missing.

To solve the above problems and realize MTSF with variable missing, forecasting models need to fully utilize historical observations of all normal variables to correct spatial-temporal dependencies during the modeling process. To this end, we propose an end-to-end framework called Graph Interpolation Attention Recursive Network (GinAR). Specifically, we use simple recursive units (SRU) based on the RNN framework as the backbone and propose two key components (interpolation attention (IA) and adaptive graph convolution (AGCN)) to replace all fully connected layers in SRU. This is done to realize end-to-end forecasting while correcting spatial-temporal dependencies. On the one hand, during the process of recursive modeling, for data at each time steps, IA first generates correspondences between normal variables and missing variables, then uses attention to restore all missing variables to plausible representations. In this way, the sequence model avoids directly mining missing variables that do not have any valuable patterns, thereby correcting temporal dependencies. On the other hand, for representations processed by IA, we use AGCN to reconstruct spatial correlations between all variables. Since all missing variables are recovered, AGCN can more accurately utilize their representations to generate a more reliable graph structure and obtain more accurate spatial correlations. In this way, GinAR mines more accurate spatial-temporal dependencies in the process of recursive modeling and effectively avoids the error accumulation problem. Thus, GinAR can implement MTSF with variable missing more accurately.
\textbf{The main contributions of this paper are as follows:}
\begin{itemize}
    \item To the best of our knowledge, this is the first work that challenges to achieve MTSF with variable missing. The proposed end-to-end framework can address the problem of error accumulation in the modeling process.
    \item To achieve this challenging task, we carefully design Graph Interpolation Attention Recursive Network, which contains two key components (interpolation attention and adaptive graph convolution). We use above components to replace all FC layers in SRU and propose the GinAR cell, aiming to correct spatial-temporal dependencies during the process of recursive modeling.
    \item We design experiments on five real-world datasets. Results show that GinAR can outperform 11 baselines on all datasets. Even when 90\% of variables are missing, it can still accurately predict the future values of all variables.
\end{itemize}

\section{Related Works}
\subsection{Spatial-Temporal Forecasting Method}
STGNNs combine the advantages of GCN ~\cite{kipf2016semi} and sequence models ~\cite{geng2022mpa} to fully mine spatial-temporal dependencies of MTS, and further improve the ability of spatial-temporal forecasting ~\cite{chengqing2023multi,liu2024rt,shao2022pre}. Li et al. ~\cite{RN58} combine gated recursive unit (GRU) and GCN to propose the diffused convolutional recurrent neural network (DCRNN) and realize MTSF. Wu et al. ~\cite{RN56} propose the graph wavenet (GWNET) by combining temporal convolutional network (TCN) and GCN. Compared with traditional methods, above two models achieves excellent results. However, these methods ignore hidden spatial correlations between variables, which limits their effectiveness ~\cite{kipf2016semi}. To further improve the ability of STGNN to mine spatial correlations, graph learning has been widely studied ~\cite{chen2023multi}. Zheng et al. ~\cite{zheng2020gman} design a spatial attention mechanism to learn attention scores by considering traffic features and variable embeddings in the graph structure. Shang et al. ~\cite{shang2021discrete} use historical observations of all variables to learn the discrete probability graph structure. Shao et al. ~\cite{shao2022decoupled} propose decoupled spatial-temporal framework and dynamic graph learning to explore spatial-temporal dependencies between variables. Although STGNNs have made significant progress in MTSF, they need to use the variable features or prior knowledge to mine spatial-temporal dependencies ~\cite{su2023novel, RN54, jiang2021dl}. However, in MTSF with variable missing, the graph structure based on prior knowledge and the graph learning based on variable features are affected by missing variables, which leads to inaccurate modeling of spatial correlations \cite{yin2023mtmgnn, deng2022multi,liang2023knowledge}. 

\subsection{Imputation Method}
Existing imputation methods include classical models ~\cite{wang2019fingerprint} and deep learning-based models ~\cite{fortuin2020gp,du2023saits,deng2021pulse}. Compared with traditional models, deep learning can analyze hidden correlations between missing and normal data and improve performance ~\cite{yoon2018gain}. Wu et al. ~\cite{RN18} combine the matrix transformation with CNN to realize the missing data imputation, but it ignores correlations between different variables, limiting its performance. Marisca et al. ~\cite{RN59} combine cross-attention and temporal attention to achieve the recovery of missing data, but they do not take full advantage of the spatial correlations between variables, which leads to inadequate data recovery. In addition to the above methods, GNN-based methods ~\cite{liang2023abslearn} combine GCN and sequence models to analyze spatial-temporal dependencies between missing data and normal data, and further recover all missing data ~\cite{wang2022multi}. Wu et al. \cite{wu2021inductive} propose inductive graph neural network to recover missing data. Compared with classical methods, the proposed model has better performance. Chen et al. ~\cite{chen2023adaptive} use the adaptive graph recursive network to realize the imputation of missing data. Experiments show that the framework combining graph convolution with recurrent neural networks can better use temporal information and spatial correlation to recover missing data. Although imputation methods can recover missing data and improve the performance of forecasting models, they often suffer from several problems: (1) Classical imputation methods ~\cite{bertsimas2021imputation} need to reconstruct both missing data and normal data, resulting in the loss of effective information. (2) Existing imputation methods ~\cite{ren2023damr,cao2018brits} require full use of temporal information to recover missing data. When the data from some variables are unavailable for a long time, existing methods introduce incorrect temporal dependencies, resulting in limited recovery performance ~\cite{chen2019traffic, liang2022reasoning}.

\section{Methodology}
\subsection{Preliminaries}

\textbf{Dependency graph.} In multivariate time series, the change of each time series depends not only on itself but also on other time series. Such a dependency can be captured by the dependency graph $G=(V, E)$. $V$ is the set of variables, and $|V|=N$. Each variable corresponds to a time series. $E$ is the set of edges. The dependency graph can be represented by an adjacency matrix: $A \in R^{N*N}$.

\textbf{Multivariate time series forecasting.} Given a historical observation tensor $X\in R^{N*H*C}$ from $H$ time slices in history, the model can predict the value $Y\in R^{N*L}$ of the nearest $L$ time steps in the future. $C$ is the number of features. The goal of MTSF is to construct a mapping function between $X\in R^{N*H*C}$ and $Y\in R^{N*L}$.

\textbf{Multivariate time series forecasting with variable missing.} Compared with MTSF, the main difference of this task is that there are some variables with whole history data missing in historical observations $X\in R^{N*H*C}$. Thus, we mask $M$ variables randomly from $N$ variables of the historical observation $X\in R^{N*H*C}$. The values of these $M$ variables are treated as 0, i.e. missing values and a new input feature $X_M\in R^{N*H*C}$ is obtained. The core goal of this task is to construct a mapping function between input $X_M\in R^{N*H*C}$ and output $Y\in R^{N*L}$.

\subsection{Overall Framework of GinAR}

\begin{figure*}
\centering
\includegraphics[width=0.8\linewidth]{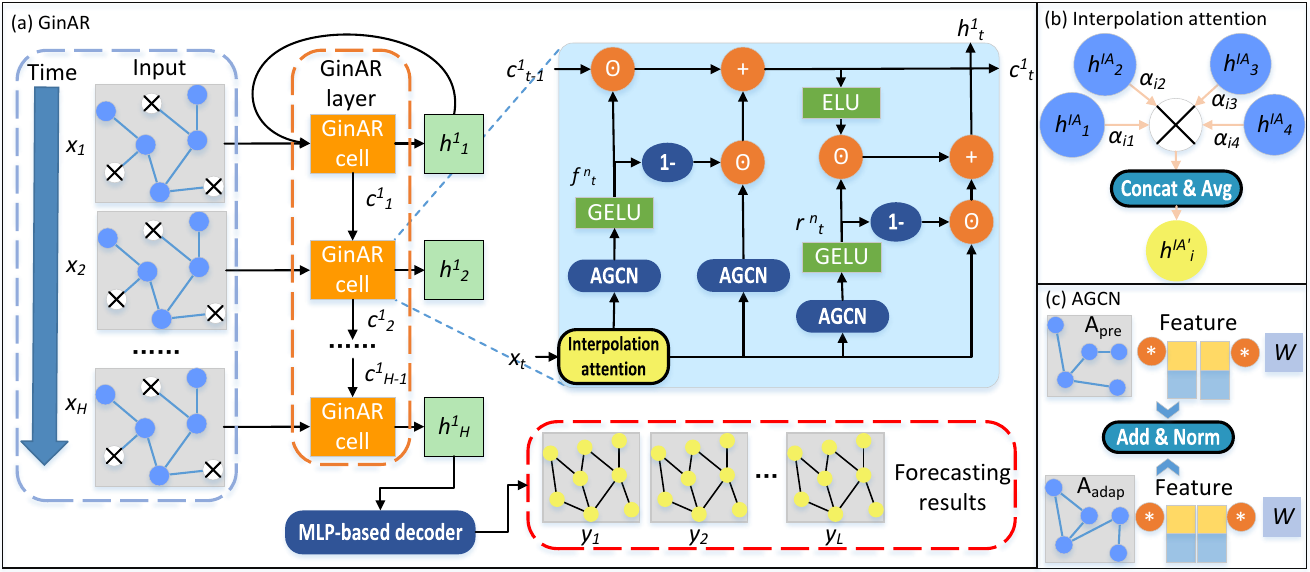}
\caption{(a) The overall framework of GinAR. The GinAR layer adopts the RNN-based sequence framework and encodes historical observations of MTS with variable missing. The MLP-based decoder is used to predict future values of all variables. (b) The specific structure of the interpolation attention. (c) The specific structure of the adaptive graph convolution. } \label{fig2}
\end{figure*}

The framework of GinAR is shown in \autoref{fig2}, which uses multiple GinAR layers as the encoder and MLP as the decoder. The GinAR layer adopts the idea of recursive modeling, and its core structure is the GinAR cell. By transmitting input features with variable missing to GinAR, it can predict future values of all variables. Next, we briefly introduce the design motivation of GinAR and the function of its components.

Firstly, we intuitively discuss the design idea for the GinAR cell. Specifically, we use IA and AGCN to replace all fully connected layers in SRU. IA can use normal variables to restore the missing variables to plausible representations, which can help the sequence model to better mine temporal dependencies of missing variables. Furthermore, for all variables processed by IA, their spatial correlations cannot be determined by a predefined graph based on prior knowledge. Therefore, we use AGCN, which introduces graph learning, to reconstruct spatial correlations between all variables. 

Then, we briefly discuss the encoder, which uses the recursive modeling framework. Specifically, at each time step $T$, the input features $x_T \in R^{N*C}$ of the current moment and the cell state $c_{T-1}$ of the previous moment are transmitted to the GinAR cell. Then, the GinAR cell outputs the cell state $c_{T}$ for the next cell and obtains the hidden feature $h_{T}$. In this way, the GinAR layer utilizes the GinAR cell to restore missing variables and reconstruct spatial correlations, while simultaneously capturing temporal dependencies through the recursive modeling framework. Besides, due to the introduction of skip connections in the GinAR cell, stacking multiple GinAR layers can capture deeper hidden information.

Finally, we discuss the decoder and the forecasting process. An important step in the forecasting process is to properly filter the hidden features obtained by the encoder. On the one hand, since the encoder takes the form of recursive modeling, the last hidden state of each GinAR layer contains all the information from the historical observation \cite{kieu2022anomaly}. On the other hand, due to the introduction of skip connections, the hidden features obtained by each GinAR layer contains different information \cite{he2016deep}. Therefore, we concatenate the last hidden state of all GinAR layers and use the concatenated tensor as the input to the decoder. Besides, we use the MLP, which is based on the direct multi-step (DMS)  forecasting strategy \cite{wu2021autoformer}, to predict future changes for all nodes. Compared with decoders based on auto-regressive \cite{cerqueira2021vest} or iterated multi-step (IMS) \cite{liu2021stochastic} forecasting, the proposed method can solve the problem of error accumulation and improve the forecasting accuracy.  

\subsection{Interpolation Attention}

For each missing variable, interpolation attention needs to select the normal variables for induction and give the corresponding weight for the selected normal variables. Thus, it contains two main steps: (1) It first generates correspondences between missing variables and normal variables. (2) Based on above correspondences, attention is used to realize the induction of missing variables. The main schematic diagram of interpolation attention is shown in \autoref{fig3}. The specific modeling steps of IA are shown below:

Step 1: First, we need to generate correspondences between missing variables and normal variables. Specifically, we initialize a diagonal matrix $I_N \in R^{N*N}$ and randomly initialize two variable-embedding matrices $E_{IA1}\in R^{N*d}$ and $E_{IA2} \in R^{d*N}$. The value of variable embedding matrix can be iterated continuously during network training. Based on following formulas, correspondences between missing variables and normal variables can be obtained:
\begin{equation}
A_{IA} = (I_N + \text{softmax}(\text{ReLU}(E_{IA1} E_{IA2})),
\end{equation}
where, \text{softmax}$(\cdot)$ is the activation function. \text{ReLU}$(\cdot)$ is the activation function. $A_{IA} \in R^{N*N} $ is a two-dimensional matrix. When the value of row $i$ and column $j$ in the $A_{IA}$ is greater than 0, it means that there is a correlation between the variable $i$ and the variable $j$. In other words, the interpolation attention can use the normal variable $j$ to recover the missing variable $i$. Based on the above variable correlation matrix $A_{IA} \in R^{N*N} $, we can obtain the set of normal variables $N (i)$ associated with the missing variable $i$.

Step 2:  Next, the missing variables $i$ are recovered by using attention mechanism and other associated normal variables $j \in N(i)$. The attention coefficient $\alpha_{ij}$ between the missing variable $i$ and normal variable $j \in N(i)$ can be calculated as follows:

\begin{equation}
\alpha_{ij} = \frac{\exp\left(\text{LeakyReLU}\left(FC(W_{j}^{IA}h_{j}^{IA})\right)\right)}{\sum_{k\in N(i)} \exp\left(\text{LeakyReLU}\left(FC(W_{k}^{IA}h_{k}^{IA}\right)\right)},
\end{equation}
where, $FC(\cdot)$ is the fully connected layer. $h_{j}^{IA}$ and $W_{j}^{IA}$ represent the features and weight of variable $j$, respectively. \text{LeakyReLU}$(\cdot)$ is the activation function. \text{exp}$(\cdot)$ stands for exponential function.

Step 3: The above attention coefficients are weighted and summed with the representations of all associated normal variables to achieve the recovery of the missing variable $i$.
\begin{equation}
h_{i}^{IA'}= \text{ReLU}(\sum_{j \in N(i)} \alpha_{ij} W_{ij}^{IA} h_{j}^{IA}),
\end{equation}

Step 4: Repeat steps 2 through 3 until all missing variables are restored. At this time, all variables have representations in the new tensor obtained by the IA method. The original input feature $X_M\in R^{N*H*C}$ can be converted to $X_M^{IA}\in R^{N*H*C'}$. (Note: This step is completed by using matrix multiplication for parallel computation.)

\begin{figure}
\centering
\includegraphics[width=0.9\linewidth]{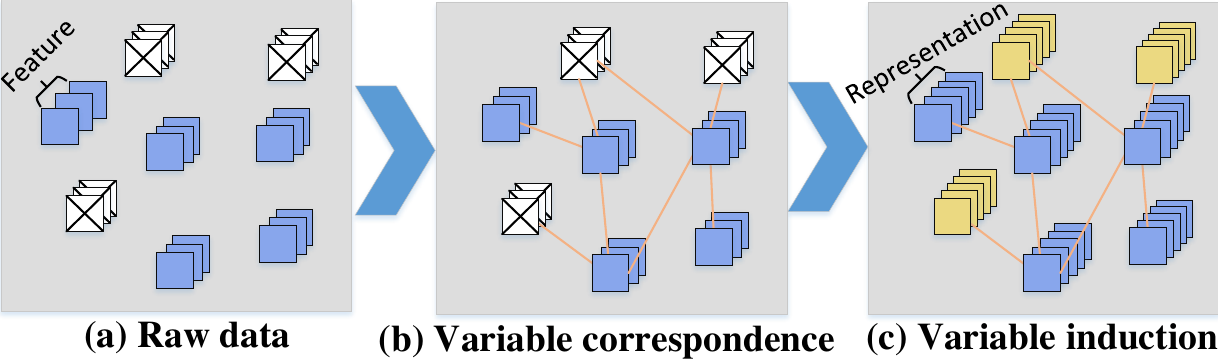}
\caption{The schematic diagram of IA. Blue represents normal variables. White represents missing variables. Yellow represents the variables after induction.} \label{fig3}
\end{figure}

\subsection{Adaptive Graph Convolution}

By introducing prior knowledge to define an adjacency matrix $A$, the predefined graph can help models establish a basic spatial correlation. However, for MTSF with missing variables, the predefined graph cannot adequately model the spatial correlation of all variables due to a large number of missing variables. To this end, we propose the data-based adaptive graph convolution, which consists of the predefined graph and the adaptive graph. 

\textbf{Predefined graph}: In this paper, distance is used to construct adjacency matrix $A$ for traffic data with road network information. For the data without road network information, the Pearson correlation coefficient ~\cite{tan2022new} is used to form the adjacency matrix $A$. The predefined graph $A_{pre}\in R^{N*N}$ for the graph convolution network is obtained by the following formula:
\begin{equation}
A_{pre}=(I_{N}+D^{-1/2} AD^{-1/2} ),
\end{equation}
where, $I_N \in R^{N*N}$ represents the diagonal matrix with value 1. $D$ is the degree matrix of $A$.

\textbf{Adaptive graph}: It needs to initialize a diagonal matrix $I_N \in R^{N*N}$ of value 1 and randomly initialize a variable-embedded matrix $E_{A}\in R^{N*d}$. The value of variable embedding matrix can be iterated continuously during neural network training. Then, based on the variable representation $X_M^{IA}\in R^{N*H*C'}$ obtained by the interpolation attention and the variable-embedded matrix $E_{A}\in R^{N*d}$, the new variable embedding $E_{n}\in R^{N*d}$ are obtained.

\begin{equation}
E_{n} = FC(\text{concat}(W_xX_M^{IA},W_eE_A)),
\end{equation}

where, $W_x$ and $W_e$ represent the weights of the variable representation $X_M^{IA}\in R^{N*H*C'}$ obtained by the interpolation attention and the variable-embedded matrix $E_{A}\in R^{N*d}$, respectively. \text{concat}$(\cdot)$ means concatenate two tensors. The adaptive graph can be obtained by the following formula:

\begin{equation}
A_{adap} = (I_N + \text{softmax}(\text{GeLU}(E_{n} E_{n}^{T})),
\end{equation}
where, $E_{n}^{T}$ represents the transpose of $E_{n}$.

\textbf{Adaptive graph convolution}: Based on the above formulas, the predefined graph and the adaptive graph can be obtained, which can reflect the spatial correlation of all variables from different perspectives. Then, we combine the adaptive graph convolution and layer normalization to fuse these graph information. The formula of the adaptive graph convolution is given as follows:

\begin{equation}
Z = F_{LN}(A_{pre} X_M^{IA} W_1 + b_1+A_{adap} X_M^{IA} W_2 + b_2),
\end{equation}
where, $X_M^{IA}$ represents the variable representation obtained by the IA. $W$ and $b$ stand for weight and bias respectively. $F_{LN}(\cdot)$ stands for the layer normalization. Through above methods, the information of adaptive graph and predefined graph is fused. 

\subsection{GinAR}

The main idea of GinAR is to integrate the proposed interpolation attention and adaptive graph convolution into the simple recursive units. 
Next, we introduce the composition of the GinAR cell and the overall modeling process of GinAR in detail.

\textbf{GinAR cell}: The GinAR cell is the most basic component of GinAR. Specifically, we introduce IA into the simple recursive unit cell to recover missing variables. Besides, we use the AGCN to replace all full connected layers in the SRU cell, enhancing the ability to correct spatial-temporal dependencies. The formula for each GinAR cell is given as follows:

\begin{equation}
x_T^{IA}=F_{IA} (x_T),
\end{equation}

\begin{equation}
\begin{aligned}
f_T=\text{GeLU}(F_{LN} (A_{pre} x_T^{IA} W_{f1}+
b_{f1}+A_{adap} x_T^{IA} W_{f2}+b_{f2})),
\end{aligned}
\end{equation}

\begin{equation}
\begin{aligned}
r_T=\text{GeLU}(F_{LN}(A_{pre} x_T^{IA} W_{r1}+ 
b_{r1} + A_{adap} x_T^{IA} W_{r2}+b_{r2})),
\end{aligned}
\end{equation}

\begin{equation}
\begin{aligned}
c_T= (1-f_T) \odot F_{LN} (A_{pre} x_T^{IA} W_{c1}+ \\
A_{adap} x_T^{IA} W_{c2} ) + f_T \odot c_{T-1},
\end{aligned}
\end{equation}

\begin{equation}
h_T=r_T \odot \text{ELU}(c_T)+(1-r_T) \odot x_T^{IA},
\end{equation}
where, $r_T$ stands for reset gate. $f_T$ stands for forget gate. $c_T$ represents the cell state of the current GinAR cell. $h_T$ is the hidden state of the current GinAR cell. $\odot$ stands for the Hadamard product. \text{GeLU}$(\cdot)$ and \text{ELU}$(\cdot)$ are activation functions. $F_{IA}(\cdot)$ stands for the interpolation attention. 

\textbf{GinAR}: The main components of GinAR include $n$ GinAR layers and an MLP-based decoder. Each GinAR layer contains multiple GinAR cells. The modeling process of GinAR is given as follows:

Step 1: The original input feature $X \in R^{N*H*C}$ is preprocessed and the input feature $X_M \in R^{N*H*C}$ for modeling is obtained. The values of the $M$ variables in the $N$ variables of the input feature $X_M \in R^{N*H*C}$ is 0.

\begin{equation}
X_M=[x_1,x_2,...,x_H],x \in R^{N*C},
\end{equation}
where, $H$ is the length of historical observation. $N$ is the number of variables. $L$ is the length of future forecasting results. $C$ stands for embedding size.

Step 2: $X_M$ is passed to the first GinAR layer. Each GinAR layer contains $H$ GinAR cells, which are used to model $x_1$ to $x_H$.

Step 3: Initialize a cell state $c_0$. $x_1$ and $c_0$ are passed to the first GinAR cell in the GinAR layer. Based on the calculation formula of GinAR cell, the hidden state $h_1^1$ of the current cell and the cell state $c_1$ are obtained. $c_1$ and $x_2$ are passed to the next GinAR cell.

Step 4: Repeat Step 3 to obtain $H$ hidden states of all GinAR cells in the first GinAR layer. These hidden states $h^1$ are used as the input features to the next GinAR layer.
\begin{equation}
h^1 = [h_1^1,h_2^1,...,h_H^1],
\end{equation}

Step 5: Repeat steps 3 to 4 until all hidden states of $n$ GinAR layers are obtained. The hidden state of the last cell in each GinAR layer is extracted. These hidden states are concatenated together as a new tensor $h_{all}^n$, which is shown as follows:

\begin{equation}
h_{all}^n = [h_H^1,h_H^2,...,h_H^n], h_{all}^n \in R^{N*C'*n},
\end{equation}

Step 6: $h_{all}^n$ is passed to the MLP-based generative decoder. And the final forecasting result $Y \in R^{N*L}$ is obtained.

\begin{equation}
Y=FC(\text{ReLU}(FC(h_{all}^n))),
\end{equation}

\section{Experimental Study}
\subsection{Experimental Design}
\textbf{Datasets.} Five real-world datasets are selected to conduct comparative experiments, including two traffic speed datasets(METR-LA and PEMS-BAY)\footnote{https://github.com/liyaguang/DCRNN}, 
two traffic flow datasets (PEMS04 and PEMS08)\footnote{https://github.com/guoshnBJTU/ASTGNN/tree/main/data} 
and an air quality dataset (China AQI)\footnote{https://quotsoft.net/air/}. 

\textbf{Baselines.} In order to fully compare and analyze the performance of the proposed GinAR, eleven existing SOTA methods are selected as the main baselines, which include forecasting models (MegaCRN ~\cite{RN52}, DSformer ~\cite{yu2023dsformer} and STID ~\cite{RN859}), 
and forecasting models with data recovery components (LGnet \cite{tang2020joint}, TriD-MAE ~\cite{zhang2023trid}, GC-VRNN ~\cite{xu2023uncovering}, and BiTGraph \cite{chen2023biased}). 
Besides, we design two-phase models (DCRNN ~\cite{RN58} + GPT4TS ~\cite{zhou2023one}, DFDGCN ~\cite{li2023dynamic} + TimesNet ~\cite{RN18} and MTGNN ~\cite{RN55} + GRIN~\cite{cinifilling}, FourierGNN ~\cite{yi2023fouriergnn}+GATGPT\cite{chen2023gatgpt}) as additional baselines to further demonstrate GinAR's effect. 

\textbf{Setting.}
\autoref{tab3} shows the main hyperparameters of the proposed model. We design experiments from the following aspects:
(1) Our code is available at this link\footnote{https://github.com/ChengqingYu/GinAR} .
(2) All datasets are uniformly divided into training sets, validation sets and test sets according to the ratio in the reference ~\cite{shao2023exploring}.
(3) We set the history/future length based on the existing work ~\cite{RN52,li2023dynamic}. The history length and future length of GinAR are both 12. Metrics are the average of 12-step forecasting results.
(4) We randomly set mask variables according to the ratio of 25\%, 50\%, 75\% and 90\%. Values of the masked variable are uniformly treated as 0. Besides, the experiment was repeated with 5 different random seeds for each missing rate. The final metrics are the mean values of repeated experiments.
(5) To ensure the fairness of experiments, we train the two-stage model in two ways: first, train the two models separately. Second, based on the reference ~\cite{xu2023uncovering}, the two models are spliced together for training. The final metrics are the optimal results.

\begin{table}
\centering
  \caption{Values of the corresponding hyperparameters for different missing rate.}
  \label{tab3}
  \begin{tabular}{cc}
\toprule
    \multirow{2}*{Config}&Values\\
    &(25\%, 50\%, 75\%, 90\%) \\
\midrule
loss function&L1 Loss\\
optimizer&Adam\\
learning rate&0.006\\
embedding size&32/32/16/16\\
variable embedding size&16/16/8/8\\
number of layers&2/2/3/3\\
dropout&0.15\\
learning rate schedule&MultiStepLR\\
clip gradient normalization&5\\
milestone&[1,15,40,70,90]\\
gamme&0.5\\
batch size&16\\
epoch&100\\
\toprule
\end{tabular}
\end{table}

\textbf{Metrics.} 
To comprehensively evaluate the forecasting performance of different models, this paper utilizes three classical metrics: Mean Absolute Error (MAE), Root Mean Square Error (RMSE), and Mean Absolute Percentage Error (MAPE) ~\cite{RN275}. 

\subsection{Main Results}

\begin{table*}
\footnotesize
\centering
\caption{Performance comparison results of all baselines and the proposed model on all datasets.}\label{tab4}
\begin{tabular}{ccccccccccccccc}
\toprule
\multirow{2}*{Datasets} & \multirow{2}*{Methods} &\multicolumn{3}{c}{Missing rate 25\%} &\multicolumn{3}{c}{Missing rate 50\%} &\multicolumn{3}{c}{Missing rate 75\%} &\multicolumn{3}{c}{Missing rate 90\%}\\
\cline{3-14} 
& & RMSE& MAPE & MAE & RMSE& MAPE & MAE  & RMSE& MAPE & MAE& RMSE& MAPE & MAE \\

\midrule
\multirow{12}*{METR-LA} 
&STID &10.31&17.85&5.05&10.73&18.88&5.35& 11.35&20.57&5.84& 12.06&23.17&6.24\\
&DSformer &7.69&11.05&4.02& 8.27&12.78&4.38&9.21& 14.59& 4.77&10.15& 17.69& 5.02\\
 &MegaCRN &7.43&10.47&3.81&7.87&11.02&3.94& 8.28&12.13&4.24& 8.72&13.54&4.58\\
&DCRNN+GPT4TS&7.41&10.72&3.78&7.91&11.61&3.98&8.16&11.93&4.15&8.31&12.18&4.29\\
&DFDGCN+TimesNet & 7.42&10.42&3.72&7.68& 11.45& 3.89&8.11&11.75&4.14&8.33&12.24&4.34\\
&MTGNN+GRIN &7.28&10.48&3.69&7.43
&11.22&3.77&8.05&11.58&4.12&8.29&12.14&4.25\\
&FourierGNN+GATGPT&
7.40& 10.87& 3.71 &7.84& 11.75 &3.82& 8.25& 12.25& 4.21& 8.37& 12.28 &4.33\\
&LGnet& 7.52& 10.97& 3.95& 8.03& 11.83& 4.17& 8.52& 13.09 &4.42& 9.15& 14.38& 4.79\\
&GC-VRNN &7.04&10.51&3.68&7.73&10.98&3.87&8.19&11.71&4.17&8.35&12.29&4.32\\
&TriD-MAE 
&7.15&10.37&3.64& 7.58&11.07&3.79&7.92&11.13&3.92&8.22&11.92&4.11\\
&BiTGraph&6.74&10.25&3.61&7.32&10.79&	3.69&7.63&11.04&3.74&8.03&11.78&3.91\\

&Proposed &\textbf{6.55}&\textbf{10.12}&\textbf{3.56}&\textbf{7.14}&\textbf{10.42}&\textbf{3.61}&\textbf{7.39}&\textbf{10.71}&\textbf{3.70}&\textbf{7.84}&\textbf{11.25}&\textbf{3.87}\\

\midrule

\multirow{12}*{PEMS-BAY} 
&STID &7.13&7.78&3.01&7.86&8.21&3.39&8.26& 9.24&3.51&8.65& 10.07&3.78\\
&DSformer 
&6.32&7.15&2.91& 6.46&7.73&3.08&8.15&9.06&3.45&9.06&10.23&3.72\\
&MegaCRN & 5.93&7.03& 2.85&7.36&7.75&3.02&7.75&8.77&3.35&8.25&9.23&3.54\\
&DCRNN+GPT4TS&5.54&6.24&2.63&6.22&6.75&2.87&7.14&7.82&3.09&7.69&8.82&3.31\\
&DFDGCN+TimesNet &5.39&6.17&2.58& 6.17&7.03&2.74&6.96&7.59&3.07&7.15&8.34&3.27\\
&MTGNN+GRIN &5.78& 5.73&2.65&6.19&6.55&2.83&7.06&7.75&3.02&7.26&8.45&3.22\\
&FourierGNN+GATGPT
 &5.21& 5.46 &2.40 &5.93& 5.96& 2.71& 6.85& 7.58& 2.98& 7.44& 9.07& 3.28 \\
&LGnet&6.02&7.19&2.88&6.74&8.15&3.14&8.08&9.18&3.43&8.92&9.83&3.67\\
&GC-VRNN &4.93&5.37&2.39&5.34&5.86&2.64&6.08&6.94&2.87&7.32&7.94&3.12\\
&TriD-MAE &5.17&5.48&2.46& 5.53&6.12&2.69&5.97&6.85&2.79&7.02&7.63&3.02\\
&BiTGraph&4.52&5.16&2.17&5.06&6.07&2.44&5.79&6.68&2.61&6.75&7.42&2.83\\
			
&Proposed &\textbf{4.34}&\textbf{4.90}&\textbf{2.10}&\textbf{4.78}&\textbf{5.88}&\textbf{2.35}&\textbf{5.48}&\textbf{6.17}&\textbf{2.54}&\textbf{6.43}&\textbf{6.94}&\textbf{2.77}\\

\midrule

\multirow{12}*{PEMS04} 
&STID &71.39 &50.88&41.96&95.09&93.63&63.73& 113.02&124.11&82.47&123.30&149.68&94.72\\
&DSformer
&50.15&20.88&32.86&51.51&21.35&33.31&54.91&23.25&37.28&59.18&25.62&40.31\\
 &MegaCRN&42.22&20.18&28.26&47.07&21.29&31.48&47.95&22.03&33.58&52.17&23.42&36.14\\
 &DCRNN+GPT4TS&40.03&17.77&25.17&41.64&18.21&26.56&43.71&19.18&28.54&46.17&21.27&31.42\\
&DFDGCN+TimesNet&39.48&17.40&24.43&41.18&18.49&26.09&42.81&19.91&28.29&45.93&21.43&30.98\\
 &MTGNN+GRIN&39.67&18.71&24.84&41.91&19.18&26.95&44.36&20.90&28.04&45.88&21.04&30.61\\
&FourierGNN+GATGPT&
 40.92& 19.61& 25.58& 42.35& 20.55& 27.31& 45.17& 22.20& 29.87& 48.83 &23.65 &32.16
\\
&LGnet&42.64&18.42&26.53&46.39&21.30&30.81&52.05&24.83&33.94&55.12&23.74&36.29\\
&GC-VRNN&39.75&16.82&23.57&41.34&17.83&26.43&43.82&18.67&27.72&45.12&20.43&30.06\\
&TriD-MAE&39.83&16.98&24.15&40.65&17.52&25.89&41.90&18.04&26.95&44.23&20.05&29.54\\
&BiTGraph&38.94&16.75&23.01&40.03&17.34&24.15&41.69&17.92&26.33&43.37&19.08&28.71\\
			
&Proposed &\textbf{38.22}&\textbf{16.45}&\textbf{22.52}&\textbf{39.02}&\textbf{17.04}&\textbf{23.78}&\textbf{41.53}&\textbf{17.58}&\textbf{25.98}&\textbf{42.82}&\textbf{18.31}&\textbf{28.20}\\

\midrule

\multirow{12}*{PEMS08} 
&STID & 58.81&31.71&32.88&79.41&51.74&49.71&101.63&74.47& 69.06&113.20&86.54&81.64\\
&DSformer &38.38&19.24&27.74&42.49&23.79&30.47&51.57&25.18& 35.21&55.34&32.61& 38.79\\
 &MegaCRN& 39.29&17.42&26.04&43.43&21.30&30.68& 48.37&21.34&32.23& 52.75&24.64&34.52\\
&DCRNN+GPT4TS &36.58&15.96&24.64& 41.79&18.73&27.96&44.82&19.79& 29.62&46.22&22.96& 31.78\\
&DFDGCN+TimesNet&36.29&15.56&24.26&39.05&19.42&25.39&42.67&20.74&28.30&45.83&21.59&30.46\\
&MTGNN+GRIN&36.65&15.17&24.08&37.64&17.41&25.78&40.51& 18.52&27.45&42.79& 21.33&29.15\\
&FourierGNN+GATGPT&
 35.54& 15.35& 23.77 &37.42& 16.86 &25.53& 39.44& 17.95& 26.89& 41.44& 20.61& 29.11
\\
&LGnet&37.54&22.18& 26.51&47.23&21.91&32.04&50.38&21.43& 33.65&52.06&25.71& 35.48\\
 &GC-VRNN&35.17&14.69&23.25&36.40&15.85&24.27&39.67&16.06&26.32&41.98&20.54&28.46\\
 &TriD-MAE &33.15&14.25&21.53&35.95&15.32&23.18&37.64&15.58&24.89&39.25&16.43&26.18\\
 &BiTGraph&31.89&14.05&20.65&35.06&14.62&22.44&36.98&15.04&23.38&39.06&16.18&25.01\\	
&Proposed &\textbf{31.34}&\textbf{13.76}&\textbf{20.41}& \textbf{34.53}&\textbf{14.21}&\textbf{22.01}&\textbf{36.04}&\textbf{14.77}&\textbf{23.10}&\textbf{38.87}&\textbf{15.82}&\textbf{24.83}\\

\midrule

\multirow{12}*{China AQI} 
&STID &31.97&45.19&18.54&34.28&48.48&20.32&36.36&55.39&22.79&40.28&62.41&25.97\\
&DSformer 
&28.22&44.81&17.73&30.35&48.09&19.06&33.22&52.05&20.63&35.72&57.31&23.17\\

&MegaCRN &28.41&33.35&15.32&29.52&35.86&16.51&33.09&48.96&19.66&35.74&53.28&22.61\\
&DCRNN+GPT4TS
&28.48&32.18&15.14&30.83&35.24&16.82&31.99&37.22&17.82&34.28&50.64&21.78\\
&DFDGCN+TimesNet 
&26.33&29.73& 14.62&29.30&32.85&15.76&30.51&38.68&17.85&33.19&51.28&21.06\\
&MTGNN+GRIN 
&27.13&33.29& 14.89&30.02&37.82&16.37&31.97&39.29&17.94&34.15&50.23&20.02\\
&FourierGNN+GATGPT&
 27.17 &32.82& 14.65& 30.77& 38.37& 16.13 &31.86& 39.90 &18.01& 34.17 &50.94 &20.18\\
&LGnet&27.76&38.92&16.02&31.03&44.95&18.39&34.09&49.63&20.36&35.54&57.18&23.24 \\
&GC-VRNN
&26.88&31.88&14.66&28.67&34.21&15.70&30.57&37.66&16.99&32.91&48.73&19.24\\

&TriD-MAE &26.18&29.09&14.51&28.96&32.94&15.74&29.84&35.76&16.79&32.68&45.76&18.04\\
&BiTGraph&25.79&28.94&13.85&27.45&31.36&14.52&29.01&33.58&15.62&31.85&38.17&17.06\\	
&Proposed &\textbf{25.51}&\textbf{28.27}&\textbf{13.72}&\textbf{26.81}&\textbf{29.56}&\textbf{14.33}&\textbf{27.96}&\textbf{31.86}&\textbf{15.39}&\textbf{30.97}&\textbf{36.30}&\textbf{16.83}\\

\toprule

\end{tabular}
\end{table*}

\autoref{tab4} gives the performance comparison results of all baselines and GinAR on five datasets (The best results are shown in \textbf{bold}). Based on results, the following conclusions can be obtained:
(1) Compared with SOTA forecasting models, all two-stage forecasting models can achieve better forecasting results. The main reason is that imputation methods use normal variables to recover missing variables, which reduces the impact of missing variables on the forecasting model. However, the error accumulation problem exists in two-stage models, which limits the performance of the downstream predictors.
(2) The forecasting models with data recovery components can work better than other baselines. On the one hand, they address the problem that one-stage models cannot handle missing data. On the other hand, they avoid the error accumulation problem of two-stage models.
(3) GinAR can achieve optimal experimental results on all datasets and all settings. Based on interpolation attention, adaptive graph convolution and RNN-based framework, the GinAR can realize missing variable recovery, spatial-temporal correlation reconstruction and end-to-end forecasting. Compared with one-stage models and two-stage models,  GinAR can avoid the problem of error accumulation and produce more accurate spatial-temporal dependencies. Therefore, GinAR can achieve better results than all baselines in MTSF with variable missing. To further evaluate the effects of each component in GinAR, we conduct ablation experiments. Besides, to demonstrate the effect of the end-to-end framework, we analyze the performance recovery effect of interpolation attention on MLP-based models.

\subsection{Ablation Experiment}

\begin{figure}
\centering
\includegraphics[width=\linewidth]{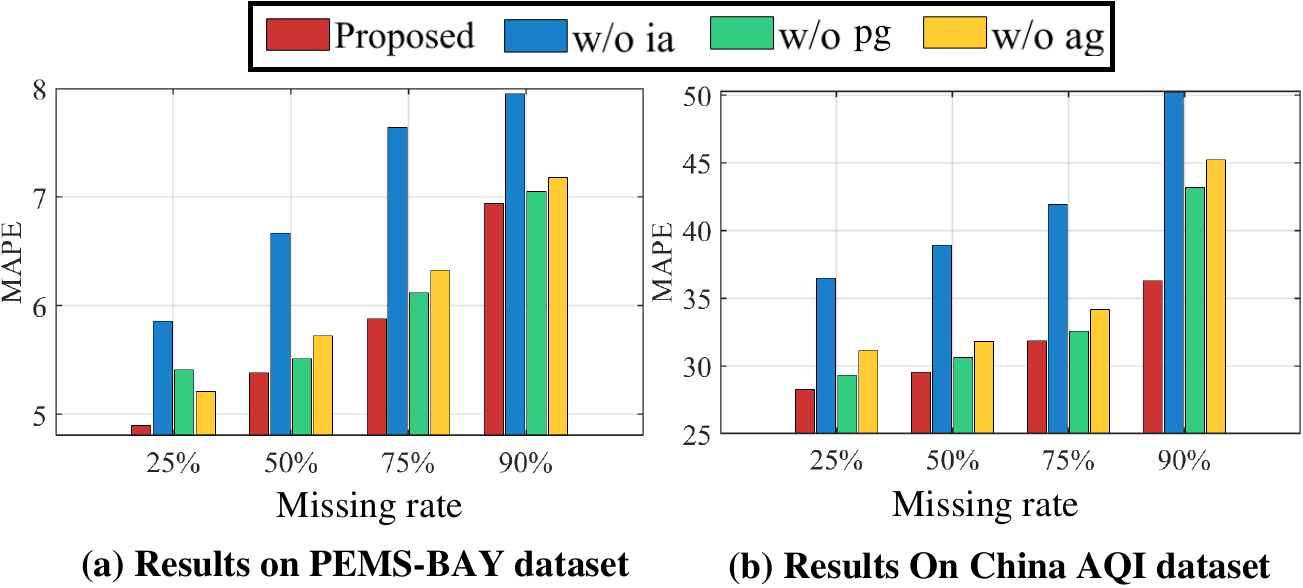}
\caption{The results of the ablation experiment.}
\label{fig5}
\end{figure}

GinAR has three important components: interpolation attention, predefined graph, and adaptive graph learning. To demonstrate the importance of these components, ablation experiments are conducted from the following three perspectives:
(1) \textbf{w/o ia}: We remove the interpolation attention. 
(2) \textbf{w/o pg}: The predefined graph is deleted. 
It means that GinAR only uses the adaptive graph to construct spatial correlations.
(3) \textbf{w/o ag}: The adaptive graph is removed. 
It means that spatial correlations are determined mainly through prior knowledge.
\autoref{fig5} shows the results of the ablation experiment.
Based on the experimental results, the following conclusions can be drawn:
(1) When the missing rate is low, deleting the predefined graph has a great impact on the forecasting result. However, when the missing rate is large, deleting the predefined graph has little impact on the result. 
(2) When the missing rate is large, deleting the adaptive graph can significantly reduce the forecasting accuracy. The main reason is that when there are more missing variables, the adaptive graph can better analyze the spatial correlation according to the characteristics of the data. Therefore, the adaptive graph plays an important role in this task. 
(3) When IA is removed, the performance of GinAR decreases significantly, proving that IA is the most important component. The main reason is that IA realizes the recovery of missing variables, which provides an important support for correcting spatial-temporal dependencies and avoiding error accumulation. 
To further analysis the effect of the IA, we compare IA with imputation methods in next section.


\subsection{Performance Evaluation of IA}
As one of the most important components proposed in this paper, it is important to further evaluate the effect of interpolation attention. In addition, it is important to further evaluate the effectiveness of the end-to-end framework. Therefore, this section compares the performance improvement effects of IA, GRIN, GATGPT, GPT4TS and TimesNet on STID. Specifically, TimesNet, GATGPT, GPT4TS and GRIN adopt the two-stage modeling framework (imputation and forecasting) to optimize the performance of STID. IA uses the end-to-end modeling framework to optimize the effects of STID. \autoref{tab5} shows the performance comparison results (MAE values) of these models (The best results are shown in boldface and the second best results are underlined). Based on the experimental results, the following conclusions can be drawn:
(1) Compared with other methods, TimesNet has minimal performance improvements to STID. The main reason is that TimesNet uses temporal information to recover missing variables, without fully analyzing correspondences between missing variables and normal variables.
(2) The proposed IA method and other imputation methods can effectively improve the forecasting effect of STID. 
(3) IA and GRIN can recover the performance of STID and obtain better forecasting results than other imputation methods. The main reason is that IA and GRIN adopt the graph-based framework to effectively reconstruct the spatial correlation between missing variables and normal variables, and then recover the missing variable data based on the normal variable. 
(4) Compared with other two-stage models, the end-to-end framework based on IA and STID can achieve good forecasting results. On the one hand, the two-stage models need to realize the feature reconstruction, and the problem of error accumulation results in the decline of forecasting accuracy. On the other hand, IA realizes adaptive induction by generating correspondences between normal variables and missing variables. 

\begin{table}
\small
\centering
\caption{MAE values of interpolation attention and other imputation methods.}\label{tab5}
\begin{tabular}{cccccc}
\toprule
\multirow{2}*{Datasets} & \multirow{2}*{Methods} &\multicolumn{4}{c}{Missing rate}  \\
\cline{3-6}
&&25\%&50\%&75\%&90\%\\

\midrule

\multirow{5}*{METR-LA} 

&STID+GPT4TS &4.05&4.32&4.76&5.02\\
&STID+TimesNet &4.54&4.64&5.13&5.44\\
&STID+GATGPT &3.96&4.19&4.39&4.57\\
&STID+GRIN &3.83&3.98&4.21&4.35\\
&STID+IA&\textbf{3.71}&\textbf{3.90}&\textbf{4.19}&\textbf{4.31}\\

\midrule

\multirow{5}*{PEMS08} 

&STID+GPT4TS & 23.84&24.91&26.79&28.51\\
&STID+TimesNet &24.55&25.24&27.70&29.69\\
&STID+GATGPT &22.49&24.18&26.43&27.76\\
&STID+GRIN &22.13&23.65&25.93&27.18\\
&STID+IA &\textbf{21.67}&\textbf{23.39}&\textbf{25.85}&\textbf{26.96}\\

\midrule

\multirow{5}*{China AQI} 

&STID+GPT4TS &15.37&16.45&17.79&19.25\\
&STID+TimesNet &15.81&17.14&18.33&19.74\\
&STID+GATGPT&14.77&15.64&16.98&18.29\\
&STID+GRIN &14.25&15.13&16.48&17.92\\
&STID+IA &\textbf{13.75}&\textbf{14.87}&\textbf{16.25}&\textbf{17.83}\\

\toprule

\end{tabular}
\end{table}

\subsection{Hyperparameter Experiment}

\begin{figure}
\centering
\includegraphics[width=0.95\linewidth]{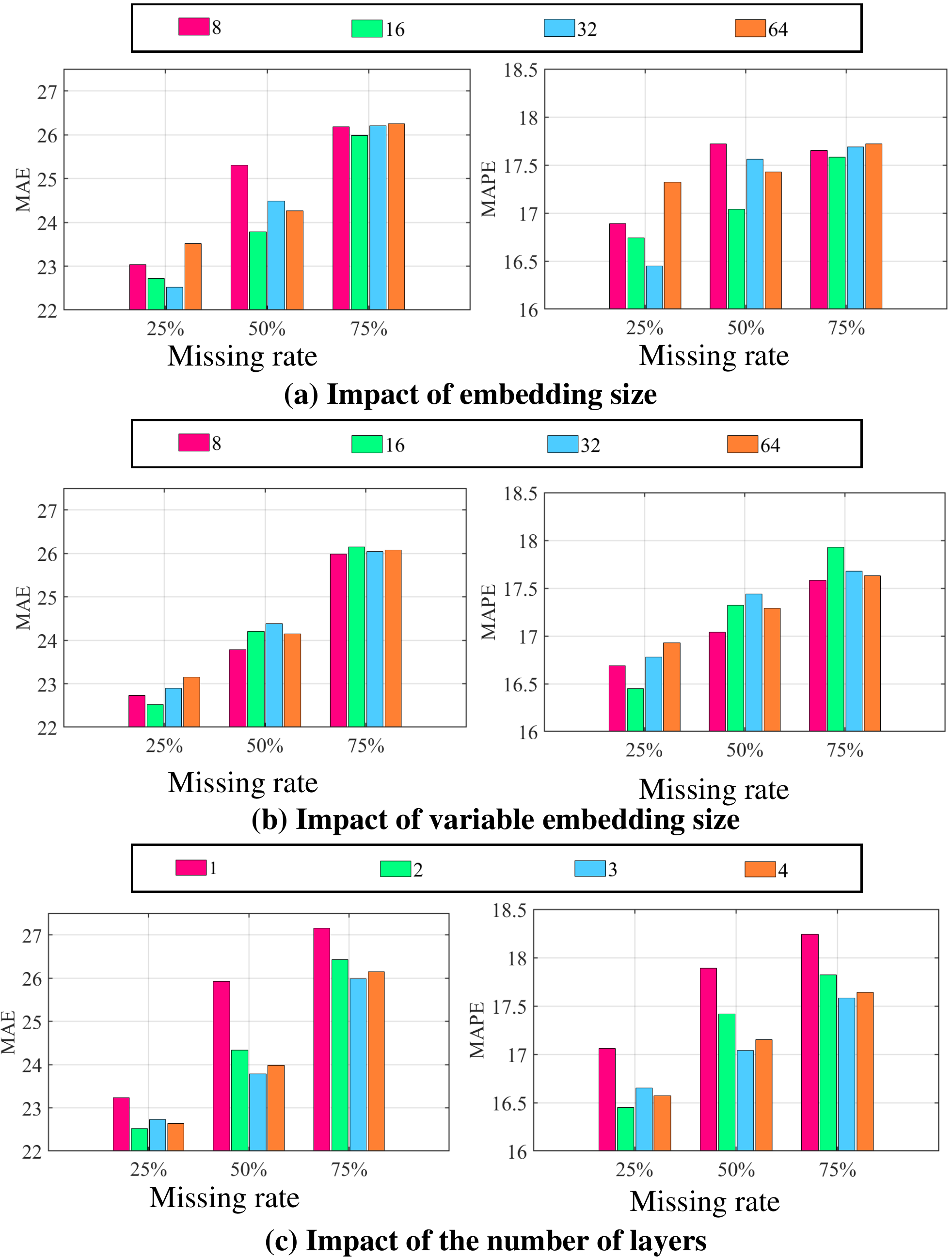}
\caption{Hyperparameter experiment results (PEMS04). }
\label{fig6}
\end{figure}

The setting of the superparameter can affect the forecasting effect of the GinAR. In this section, we evaluate the influence of three main hyperparameters on the experimental results, including embedding size, variable embedding size, and number of layers. \autoref{fig6} shows the influence of different hyperparameters on the experimental results (PEMS04 dataset). Based on the experimental results, the following conclusions can be obtained:
(1) The variable embedding size has the least influence on the forecasting results. It proves that the adaptive graph with good performance can be generated without a large number of parameters. 
(2) The embedding size can be increased appropriately when the missing rate is small. And the embedding size cannot be too large when the missing rate is large. The main reason is that when the missing rate is large, the large embedding size can lead to overfitting, thus affecting the forecasting accuracy. 
(3) The number of layers has the greatest influence on the forecasting result. Too few layers can not adequately mine and analyze the data. Too many layers can lead to problems such as overfitting. Therefore, the best forecasting results is achieved when the number of layers is set to 2 or 3.

\subsection{Visualization}

\begin{figure}[h]
\centering
\includegraphics[width=\linewidth]{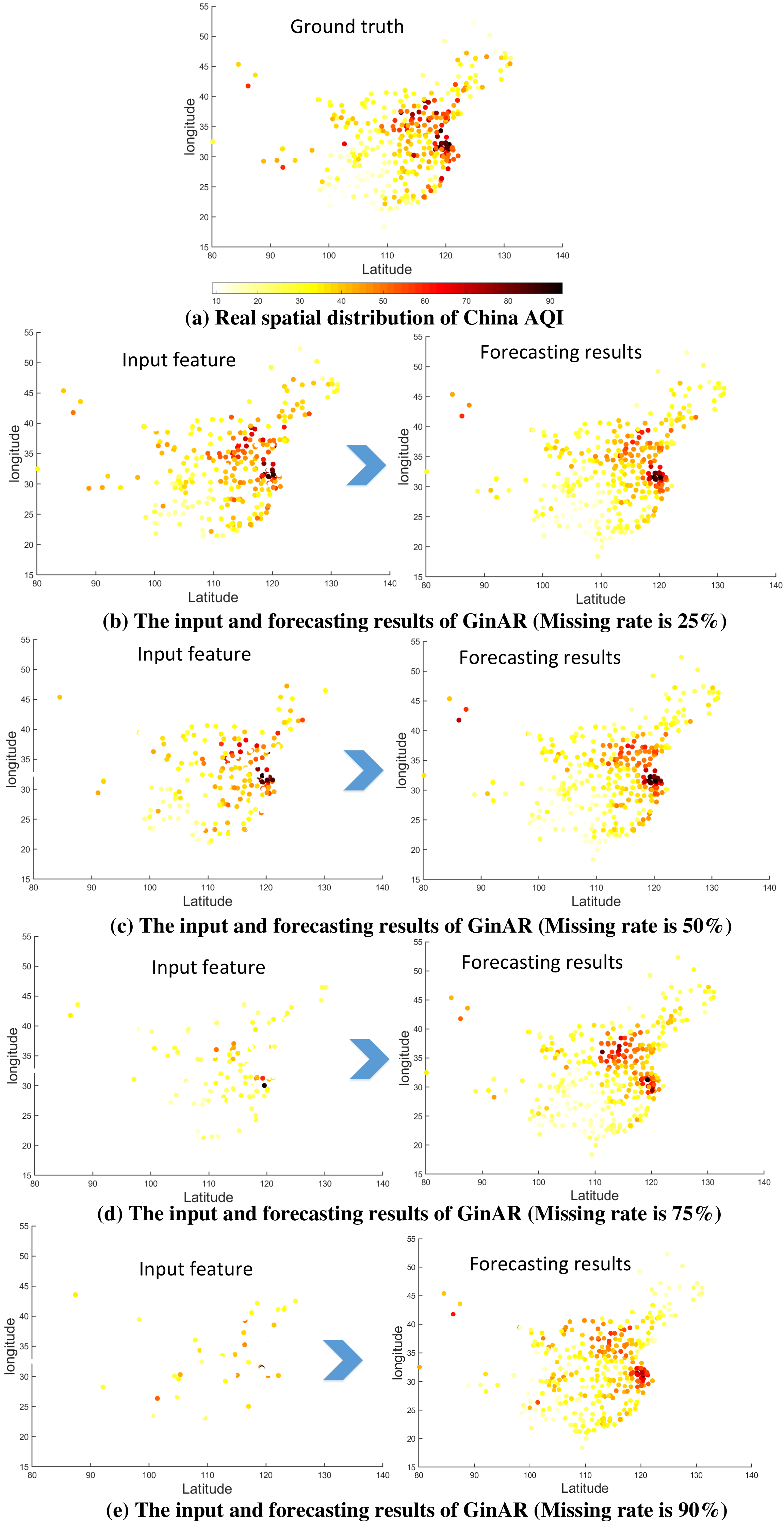}
\caption{Visualization of the input features and forecasting results of GinAR on different missing rates (China AQI dataset). As the missing rate increases, the input features become more and more sparse. However, the forecasting performence of GinAR does not deteriorate significantly.}
\label{fig8}
\end{figure}

To prove that GinAR can effectively predict the future values of all variables, this section visualizes the forecasting results from the spatial dimension. \autoref{fig8} gives the visualization of the input features and forecasting results of GinAR on different missing rates (China AQI dataset). Based on the visualization results, we can get the following conclusions:
(1)  As shown in \autoref{fig8} (a), \autoref{fig8} (b) and \autoref{fig8} (c), GinAR can accurately predict the AQI value of all variables when the missing rate is not particularly large. 
(2) As shown in \autoref{fig8} (a), \autoref{fig8} (d) and \autoref{fig8} (e), even though the number of normal variables is very sparse, GinAR can still accurately predict the spatial distribution of the AQI data. 
(3) GinAR can make full use of normal variables to realize accurate spatial-temporal forecasting for all variables. The visualization results can further demonstrate the practical value of GinAR. 

\section{Conclusion and Future Work}
In this paper, we try to address a new challenging task: MTSF with variable missing. In this task, to solve the problems of producing incorrect spatial-temporal dependencies and error accumulation in existing models, we carefully design two key components (Interpolation Attention and Adaptive Graph Convolution) and use them to replace all fully connected layers in the simple recursive unit. In this way, we propose the Graph Interpolation Attention Recursive Network based on the end-to-end framework, which can simultaneously recover all missing variables, correct spatial-temporal dependencies, and predict the future values of all variables. Experimental results on five real-world datasets demonstrate the practical value of our model, and even when only 10\% of variables are normal, it can predict the future values of all the variables. In the future, we will optimize the efficiency of GinAR and work on datasets with larger spatial dimensions and more complex spatial correlations.

\begin{acks}
This work is  supported by NSFC No. 62372430, NSFC No. 62206266 and the Youth Innovation Promotion Association CAS No.2023112.
\end{acks}


\clearpage
\bibliographystyle{ACM-Reference-Format}
\bibliography{sample-base}


\begin{thebibliography}{82}


\ifx \showCODEN    \undefined \def \showCODEN     #1{\unskip}     \fi
\ifx \showDOI      \undefined \def \showDOI       #1{#1}\fi
\ifx \showISBNx    \undefined \def \showISBNx     #1{\unskip}     \fi
\ifx \showISBNxiii \undefined \def \showISBNxiii  #1{\unskip}     \fi
\ifx \showISSN     \undefined \def \showISSN      #1{\unskip}     \fi
\ifx \showLCCN     \undefined \def \showLCCN      #1{\unskip}     \fi
\ifx \shownote     \undefined \def \shownote      #1{#1}          \fi
\ifx \showarticletitle \undefined \def \showarticletitle #1{#1}   \fi
\ifx \showURL      \undefined \def \showURL       {\relax}        \fi
\providecommand\bibfield[2]{#2}
\providecommand\bibinfo[2]{#2}
\providecommand\natexlab[1]{#1}
\providecommand\showeprint[2][]{arXiv:#2}

\bibitem[Bertsimas et~al\mbox{.}(2021)]%
        {bertsimas2021imputation}
\bibfield{author}{\bibinfo{person}{Dimitris Bertsimas}, \bibinfo{person}{Agni Orfanoudaki}, {and} \bibinfo{person}{Colin Pawlowski}.} \bibinfo{year}{2021}\natexlab{}.
\newblock \showarticletitle{Imputation of clinical covariates in time series}.
\newblock \bibinfo{journal}{\emph{Machine Learning}}  \bibinfo{volume}{110} (\bibinfo{year}{2021}), \bibinfo{pages}{185--248}.
\newblock


\bibitem[Bl{\'a}zquez-Garc{\'\i}a et~al\mbox{.}(2023)]%
        {blazquez2023selective}
\bibfield{author}{\bibinfo{person}{Ane Bl{\'a}zquez-Garc{\'\i}a}, \bibinfo{person}{Kristoffer Wickstr{\o}m}, \bibinfo{person}{Shujian Yu}, \bibinfo{person}{Karl~{\O}yvind Mikalsen}, \bibinfo{person}{Ahcene Boubekki}, \bibinfo{person}{Angel Conde}, \bibinfo{person}{Usue Mori}, \bibinfo{person}{Robert Jenssen}, {and} \bibinfo{person}{Jose~A Lozano}.} \bibinfo{year}{2023}\natexlab{}.
\newblock \showarticletitle{Selective imputation for multivariate time series datasets with missing values}.
\newblock \bibinfo{journal}{\emph{IEEE Transactions on Knowledge and Data Engineering}} (\bibinfo{year}{2023}).
\newblock


\bibitem[Cao et~al\mbox{.}(2020)]%
        {cao2020spectral}
\bibfield{author}{\bibinfo{person}{Defu Cao}, \bibinfo{person}{Yujing Wang}, \bibinfo{person}{Juanyong Duan}, \bibinfo{person}{Ce Zhang}, \bibinfo{person}{Xia Zhu}, \bibinfo{person}{Congrui Huang}, \bibinfo{person}{Yunhai Tong}, \bibinfo{person}{Bixiong Xu}, \bibinfo{person}{Jing Bai}, \bibinfo{person}{Jie Tong}, {et~al\mbox{.}}} \bibinfo{year}{2020}\natexlab{}.
\newblock \showarticletitle{Spectral temporal graph neural network for multivariate time-series forecasting}.
\newblock \bibinfo{journal}{\emph{Advances in neural information processing systems}}  \bibinfo{volume}{33} (\bibinfo{year}{2020}), \bibinfo{pages}{17766--17778}.
\newblock


\bibitem[Cao et~al\mbox{.}(2018)]%
        {cao2018brits}
\bibfield{author}{\bibinfo{person}{Wei Cao}, \bibinfo{person}{Dong Wang}, \bibinfo{person}{Jian Li}, \bibinfo{person}{Hao Zhou}, \bibinfo{person}{Lei Li}, {and} \bibinfo{person}{Yitan Li}.} \bibinfo{year}{2018}\natexlab{}.
\newblock \showarticletitle{Brits: Bidirectional recurrent imputation for time series}.
\newblock \bibinfo{journal}{\emph{Advances in neural information processing systems}}  \bibinfo{volume}{31} (\bibinfo{year}{2018}).
\newblock


\bibitem[Cerqueira et~al\mbox{.}(2021)]%
        {cerqueira2021vest}
\bibfield{author}{\bibinfo{person}{Vitor Cerqueira}, \bibinfo{person}{Nuno Moniz}, {and} \bibinfo{person}{Carlos Soares}.} \bibinfo{year}{2021}\natexlab{}.
\newblock \showarticletitle{Vest: Automatic feature engineering for forecasting}.
\newblock \bibinfo{journal}{\emph{Machine Learning}} (\bibinfo{year}{2021}), \bibinfo{pages}{1--23}.
\newblock


\bibitem[Chauhan et~al\mbox{.}(2022)]%
        {chauhan2022multi}
\bibfield{author}{\bibinfo{person}{Jatin Chauhan}, \bibinfo{person}{Aravindan Raghuveer}, \bibinfo{person}{Rishi Saket}, \bibinfo{person}{Jay Nandy}, {and} \bibinfo{person}{Balaraman Ravindran}.} \bibinfo{year}{2022}\natexlab{}.
\newblock \showarticletitle{Multi-Variate Time Series Forecasting on Variable Subsets}. In \bibinfo{booktitle}{\emph{Proceedings of the 28th ACM SIGKDD Conference on Knowledge Discovery and Data Mining}}. \bibinfo{pages}{76--86}.
\newblock


\bibitem[Chen et~al\mbox{.}(2023b)]%
        {chen2023multi}
\bibfield{author}{\bibinfo{person}{Ling Chen}, \bibinfo{person}{Donghui Chen}, \bibinfo{person}{Zongjiang Shang}, \bibinfo{person}{Binqing Wu}, \bibinfo{person}{Cen Zheng}, \bibinfo{person}{Bo Wen}, {and} \bibinfo{person}{Wei Zhang}.} \bibinfo{year}{2023}\natexlab{b}.
\newblock \showarticletitle{Multi-scale adaptive graph neural network for multivariate time series forecasting}.
\newblock \bibinfo{journal}{\emph{IEEE Transactions on Knowledge and Data Engineering}} (\bibinfo{year}{2023}).
\newblock


\bibitem[Chen et~al\mbox{.}(2021)]%
        {chen2021low}
\bibfield{author}{\bibinfo{person}{Xinyu Chen}, \bibinfo{person}{Mengying Lei}, \bibinfo{person}{Nicolas Saunier}, {and} \bibinfo{person}{Lijun Sun}.} \bibinfo{year}{2021}\natexlab{}.
\newblock \showarticletitle{Low-rank autoregressive tensor completion for spatiotemporal traffic data imputation}.
\newblock \bibinfo{journal}{\emph{IEEE Transactions on Intelligent Transportation Systems}} \bibinfo{volume}{23}, \bibinfo{number}{8} (\bibinfo{year}{2021}), \bibinfo{pages}{12301--12310}.
\newblock


\bibitem[Chen et~al\mbox{.}(2023c)]%
        {chen2023biased}
\bibfield{author}{\bibinfo{person}{Xiaodan Chen}, \bibinfo{person}{Xiucheng Li}, \bibinfo{person}{Bo Liu}, {and} \bibinfo{person}{Zhijun Li}.} \bibinfo{year}{2023}\natexlab{c}.
\newblock \showarticletitle{Biased Temporal Convolution Graph Network for Time Series Forecasting with Missing Values.}. In \bibinfo{booktitle}{\emph{The Twelfth International Conference on Learning Representations}}.
\newblock


\bibitem[Chen et~al\mbox{.}(2023a)]%
        {chen2023higher}
\bibfield{author}{\bibinfo{person}{Yuzhou Chen}, \bibinfo{person}{Sotiris Batsakis}, {and} \bibinfo{person}{H~Vincent Poor}.} \bibinfo{year}{2023}\natexlab{a}.
\newblock \showarticletitle{Higher-Order Spatio-Temporal Neural Networks for Covid-19 Forecasting}. In \bibinfo{booktitle}{\emph{ICASSP 2023-2023 IEEE International Conference on Acoustics, Speech and Signal Processing (ICASSP)}}. IEEE, \bibinfo{pages}{1--5}.
\newblock


\bibitem[Chen and Chen(2022)]%
        {chen2022novel}
\bibfield{author}{\bibinfo{person}{Yong Chen} {and} \bibinfo{person}{Xiqun~Michael Chen}.} \bibinfo{year}{2022}\natexlab{}.
\newblock \showarticletitle{A novel reinforced dynamic graph convolutional network model with data imputation for network-wide traffic flow prediction}.
\newblock \bibinfo{journal}{\emph{Transportation Research Part C: Emerging Technologies}}  \bibinfo{volume}{143} (\bibinfo{year}{2022}), \bibinfo{pages}{103820}.
\newblock


\bibitem[Chen et~al\mbox{.}(2023d)]%
        {chen2023adaptive}
\bibfield{author}{\bibinfo{person}{Yakun Chen}, \bibinfo{person}{Zihao Li}, \bibinfo{person}{Chao Yang}, \bibinfo{person}{Xianzhi Wang}, \bibinfo{person}{Guodong Long}, {and} \bibinfo{person}{Guandong Xu}.} \bibinfo{year}{2023}\natexlab{d}.
\newblock \showarticletitle{Adaptive graph recurrent network for multivariate time series imputation}. In \bibinfo{booktitle}{\emph{Neural Information Processing: 29th International Conference, ICONIP 2022, Virtual Event, November 22--26, 2022, Proceedings, Part V}}. Springer, \bibinfo{pages}{64--73}.
\newblock


\bibitem[Chen et~al\mbox{.}(2019)]%
        {chen2019traffic}
\bibfield{author}{\bibinfo{person}{Yuanyuan Chen}, \bibinfo{person}{Yisheng Lv}, {and} \bibinfo{person}{Fei-Yue Wang}.} \bibinfo{year}{2019}\natexlab{}.
\newblock \showarticletitle{Traffic flow imputation using parallel data and generative adversarial networks}.
\newblock \bibinfo{journal}{\emph{IEEE Transactions on Intelligent Transportation Systems}} \bibinfo{volume}{21}, \bibinfo{number}{4} (\bibinfo{year}{2019}), \bibinfo{pages}{1624--1630}.
\newblock


\bibitem[Chen et~al\mbox{.}(2023e)]%
        {chen2023gatgpt}
\bibfield{author}{\bibinfo{person}{Yakun Chen}, \bibinfo{person}{Xianzhi Wang}, {and} \bibinfo{person}{Guandong Xu}.} \bibinfo{year}{2023}\natexlab{e}.
\newblock \showarticletitle{Gatgpt: A pre-trained large language model with graph attention network for spatiotemporal imputation}.
\newblock \bibinfo{journal}{\emph{arXiv preprint arXiv:2311.14332}} (\bibinfo{year}{2023}).
\newblock


\bibitem[Chengqing et~al\mbox{.}(2023)]%
        {chengqing2023multi}
\bibfield{author}{\bibinfo{person}{Yu Chengqing}, \bibinfo{person}{Yan Guangxi}, \bibinfo{person}{Yu Chengming}, \bibinfo{person}{Zhang Yu}, {and} \bibinfo{person}{Mi Xiwei}.} \bibinfo{year}{2023}\natexlab{}.
\newblock \showarticletitle{A multi-factor driven spatiotemporal wind power prediction model based on ensemble deep graph attention reinforcement learning networks}.
\newblock \bibinfo{journal}{\emph{Energy}}  \bibinfo{volume}{263} (\bibinfo{year}{2023}), \bibinfo{pages}{126034}.
\newblock


\bibitem[Chowdhury et~al\mbox{.}(2022)]%
        {chowdhury2022tarnet}
\bibfield{author}{\bibinfo{person}{Ranak~Roy Chowdhury}, \bibinfo{person}{Xiyuan Zhang}, \bibinfo{person}{Jingbo Shang}, \bibinfo{person}{Rajesh~K Gupta}, {and} \bibinfo{person}{Dezhi Hong}.} \bibinfo{year}{2022}\natexlab{}.
\newblock \showarticletitle{Tarnet: Task-aware reconstruction for time-series transformer}. In \bibinfo{booktitle}{\emph{Proceedings of the 28th ACM SIGKDD Conference on Knowledge Discovery and Data Mining}}. \bibinfo{pages}{212--220}.
\newblock


\bibitem[Cini et~al\mbox{.}(2022)]%
        {cinifilling}
\bibfield{author}{\bibinfo{person}{Andrea Cini}, \bibinfo{person}{Ivan Marisca}, {and} \bibinfo{person}{Cesare Alippi}.} \bibinfo{year}{2022}\natexlab{}.
\newblock \showarticletitle{Filling the G\_ap\_s: Multivariate Time Series Imputation by Graph Neural Networks}. In \bibinfo{booktitle}{\emph{International Conference on Learning Representations}}.
\newblock


\bibitem[Cirstea et~al\mbox{.}(2022)]%
        {cirstea2022towards}
\bibfield{author}{\bibinfo{person}{Razvan-Gabriel Cirstea}, \bibinfo{person}{Bin Yang}, \bibinfo{person}{Chenjuan Guo}, \bibinfo{person}{Tung Kieu}, {and} \bibinfo{person}{Shirui Pan}.} \bibinfo{year}{2022}\natexlab{}.
\newblock \showarticletitle{Towards spatio-temporal aware traffic time series forecasting}. In \bibinfo{booktitle}{\emph{2022 IEEE 38th International Conference on Data Engineering (ICDE)}}. IEEE, \bibinfo{pages}{2900--2913}.
\newblock


\bibitem[Deng et~al\mbox{.}(2021a)]%
        {deng2021pulse}
\bibfield{author}{\bibinfo{person}{Jinliang Deng}, \bibinfo{person}{Xiusi Chen}, \bibinfo{person}{Zipei Fan}, \bibinfo{person}{Renhe Jiang}, \bibinfo{person}{Xuan Song}, {and} \bibinfo{person}{Ivor~W Tsang}.} \bibinfo{year}{2021}\natexlab{a}.
\newblock \showarticletitle{The pulse of urban transport: Exploring the co-evolving pattern for spatio-temporal forecasting}.
\newblock \bibinfo{journal}{\emph{ACM Transactions on Knowledge Discovery from Data (TKDD)}} \bibinfo{volume}{15}, \bibinfo{number}{6} (\bibinfo{year}{2021}), \bibinfo{pages}{1--25}.
\newblock


\bibitem[Deng et~al\mbox{.}(2021b)]%
        {RN54}
\bibfield{author}{\bibinfo{person}{Jinliang Deng}, \bibinfo{person}{Xiusi Chen}, \bibinfo{person}{Renhe Jiang}, \bibinfo{person}{Xuan Song}, {and} \bibinfo{person}{Ivor~W Tsang}.} \bibinfo{year}{2021}\natexlab{b}.
\newblock \showarticletitle{St-norm: Spatial and temporal normalization for multi-variate time series forecasting}. In \bibinfo{booktitle}{\emph{Proceedings of the 27th ACM SIGKDD conference on knowledge discovery \& data mining}}. \bibinfo{pages}{269--278}.
\newblock


\bibitem[Deng et~al\mbox{.}(2022)]%
        {deng2022multi}
\bibfield{author}{\bibinfo{person}{Jinliang Deng}, \bibinfo{person}{Xiusi Chen}, \bibinfo{person}{Renhe Jiang}, \bibinfo{person}{Xuan Song}, {and} \bibinfo{person}{Ivor~W Tsang}.} \bibinfo{year}{2022}\natexlab{}.
\newblock \showarticletitle{A multi-view multi-task learning framework for multi-variate time series forecasting}.
\newblock \bibinfo{journal}{\emph{IEEE Transactions on Knowledge and Data Engineering}} (\bibinfo{year}{2022}).
\newblock


\bibitem[Deng et~al\mbox{.}(2024)]%
        {deng2024disentangling}
\bibfield{author}{\bibinfo{person}{Jinliang Deng}, \bibinfo{person}{Xiusi Chen}, \bibinfo{person}{Renhe Jiang}, \bibinfo{person}{Du Yin}, \bibinfo{person}{Yi Yang}, \bibinfo{person}{Xuan Song}, {and} \bibinfo{person}{Ivor~W Tsang}.} \bibinfo{year}{2024}\natexlab{}.
\newblock \showarticletitle{Disentangling Structured Components: Towards Adaptive, Interpretable and Scalable Time Series Forecasting}.
\newblock \bibinfo{journal}{\emph{IEEE Transactions on Knowledge and Data Engineering}} (\bibinfo{year}{2024}).
\newblock


\bibitem[Du et~al\mbox{.}(2023)]%
        {du2023saits}
\bibfield{author}{\bibinfo{person}{Wenjie Du}, \bibinfo{person}{David C{\^o}t{\'e}}, {and} \bibinfo{person}{Yan Liu}.} \bibinfo{year}{2023}\natexlab{}.
\newblock \showarticletitle{Saits: Self-attention-based imputation for time series}.
\newblock \bibinfo{journal}{\emph{Expert Systems with Applications}}  \bibinfo{volume}{219} (\bibinfo{year}{2023}), \bibinfo{pages}{119619}.
\newblock


\bibitem[Fortuin et~al\mbox{.}(2020)]%
        {fortuin2020gp}
\bibfield{author}{\bibinfo{person}{Vincent Fortuin}, \bibinfo{person}{Dmitry Baranchuk}, \bibinfo{person}{Gunnar R{\"a}tsch}, {and} \bibinfo{person}{Stephan Mandt}.} \bibinfo{year}{2020}\natexlab{}.
\newblock \showarticletitle{Gp-vae: Deep probabilistic time series imputation}. In \bibinfo{booktitle}{\emph{International conference on artificial intelligence and statistics}}. PMLR, \bibinfo{pages}{1651--1661}.
\newblock


\bibitem[Geng et~al\mbox{.}(2022)]%
        {geng2022mpa}
\bibfield{author}{\bibinfo{person}{Jingxuan Geng}, \bibinfo{person}{Chunhua Yang}, \bibinfo{person}{Yonggang Li}, \bibinfo{person}{Lijuan Lan}, {and} \bibinfo{person}{Qiwu Luo}.} \bibinfo{year}{2022}\natexlab{}.
\newblock \showarticletitle{MPA-RNN: a novel attention-based recurrent neural networks for total nitrogen prediction}.
\newblock \bibinfo{journal}{\emph{IEEE Transactions on Industrial Informatics}} \bibinfo{volume}{18}, \bibinfo{number}{10} (\bibinfo{year}{2022}), \bibinfo{pages}{6516--6525}.
\newblock


\bibitem[He et~al\mbox{.}(2016)]%
        {he2016deep}
\bibfield{author}{\bibinfo{person}{Kaiming He}, \bibinfo{person}{Xiangyu Zhang}, \bibinfo{person}{Shaoqing Ren}, {and} \bibinfo{person}{Jian Sun}.} \bibinfo{year}{2016}\natexlab{}.
\newblock \showarticletitle{Deep residual learning for image recognition}. In \bibinfo{booktitle}{\emph{Proceedings of the IEEE conference on computer vision and pattern recognition}}. \bibinfo{pages}{770--778}.
\newblock


\bibitem[Hu et~al\mbox{.}(2022)]%
        {hu2022mgcn}
\bibfield{author}{\bibinfo{person}{Jia Hu}, \bibinfo{person}{Xianghong Lin}, {and} \bibinfo{person}{Chu Wang}.} \bibinfo{year}{2022}\natexlab{}.
\newblock \showarticletitle{MGCN: Dynamic Spatio-Temporal Multi-Graph Convolutional Neural Network}. In \bibinfo{booktitle}{\emph{2022 International Joint Conference on Neural Networks (IJCNN)}}. IEEE, \bibinfo{pages}{1--9}.
\newblock


\bibitem[Ivan et~al\mbox{.}(2022)]%
        {RN59}
\bibfield{author}{\bibinfo{person}{Marisca Ivan}, \bibinfo{person}{Cini Andrea}, {and} \bibinfo{person}{Cesare Alippi}.} \bibinfo{year}{2022}\natexlab{}.
\newblock \showarticletitle{Learning to Reconstruct Missing Data from Spatiotemporal Graphs with Sparse Observations}. In \bibinfo{booktitle}{\emph{36th Conference on Neural Information Processing Systems (NeurIPS 2022)}}. \bibinfo{pages}{1--17}.
\newblock


\bibitem[Jiang et~al\mbox{.}(2023)]%
        {RN52}
\bibfield{author}{\bibinfo{person}{Renhe Jiang}, \bibinfo{person}{Zhaonan Wang}, \bibinfo{person}{Jiawei Yong}, \bibinfo{person}{Puneet Jeph}, \bibinfo{person}{Quanjun Chen}, \bibinfo{person}{Yasumasa Kobayashi}, \bibinfo{person}{Xuan Song}, \bibinfo{person}{Shintaro Fukushima}, {and} \bibinfo{person}{Toyotaro Suzumura}.} \bibinfo{year}{2023}\natexlab{}.
\newblock \showarticletitle{Spatio-temporal meta-graph learning for traffic forecasting}. In \bibinfo{booktitle}{\emph{Proceedings of the AAAI Conference on Artificial Intelligence}}, Vol.~\bibinfo{volume}{37}. \bibinfo{pages}{8078--8086}.
\newblock


\bibitem[Jiang et~al\mbox{.}(2021)]%
        {jiang2021dl}
\bibfield{author}{\bibinfo{person}{Renhe Jiang}, \bibinfo{person}{Du Yin}, \bibinfo{person}{Zhaonan Wang}, \bibinfo{person}{Yizhuo Wang}, \bibinfo{person}{Jiewen Deng}, \bibinfo{person}{Hangchen Liu}, \bibinfo{person}{Zekun Cai}, \bibinfo{person}{Jinliang Deng}, \bibinfo{person}{Xuan Song}, {and} \bibinfo{person}{Ryosuke Shibasaki}.} \bibinfo{year}{2021}\natexlab{}.
\newblock \showarticletitle{Dl-traff: Survey and benchmark of deep learning models for urban traffic prediction}. In \bibinfo{booktitle}{\emph{Proceedings of the 30th ACM international conference on information \& knowledge management}}. \bibinfo{pages}{4515--4525}.
\newblock


\bibitem[Kieu et~al\mbox{.}(2022)]%
        {kieu2022anomaly}
\bibfield{author}{\bibinfo{person}{Tung Kieu}, \bibinfo{person}{Bin Yang}, \bibinfo{person}{Chenjuan Guo}, \bibinfo{person}{Razvan-Gabriel Cirstea}, \bibinfo{person}{Yan Zhao}, \bibinfo{person}{Yale Song}, {and} \bibinfo{person}{Christian~S Jensen}.} \bibinfo{year}{2022}\natexlab{}.
\newblock \showarticletitle{Anomaly detection in time series with robust variational quasi-recurrent autoencoders}. In \bibinfo{booktitle}{\emph{2022 IEEE 38th International Conference on Data Engineering (ICDE)}}. IEEE, \bibinfo{pages}{1342--1354}.
\newblock


\bibitem[Kipf and Welling(2016)]%
        {kipf2016semi}
\bibfield{author}{\bibinfo{person}{Thomas~N Kipf} {and} \bibinfo{person}{Max Welling}.} \bibinfo{year}{2016}\natexlab{}.
\newblock \showarticletitle{Semi-Supervised Classification with Graph Convolutional Networks}. In \bibinfo{booktitle}{\emph{International Conference on Learning Representations}}.
\newblock


\bibitem[Li et~al\mbox{.}(2023b)]%
        {li2023multivariate}
\bibfield{author}{\bibinfo{person}{Jinlong Li}, \bibinfo{person}{Pan Wu}, \bibinfo{person}{Hengcong Guo}, \bibinfo{person}{Ruonan Li}, \bibinfo{person}{Guilin Li}, {and} \bibinfo{person}{Lunhui Xu}.} \bibinfo{year}{2023}\natexlab{b}.
\newblock \showarticletitle{Multivariate Transfer Passenger Flow Forecasting with Data Imputation by Joint Deep Learning and Matrix Factorization}.
\newblock \bibinfo{journal}{\emph{Applied Sciences}} \bibinfo{volume}{13}, \bibinfo{number}{9} (\bibinfo{year}{2023}), \bibinfo{pages}{5625}.
\newblock


\bibitem[Li et~al\mbox{.}(2023a)]%
        {li2023dynamic}
\bibfield{author}{\bibinfo{person}{Yujie Li}, \bibinfo{person}{Zezhi Shao}, \bibinfo{person}{Yongjun Xu}, \bibinfo{person}{Qiang Qiu}, \bibinfo{person}{Zhaogang Cao}, {and} \bibinfo{person}{Fei Wang}.} \bibinfo{year}{2023}\natexlab{a}.
\newblock \showarticletitle{Dynamic Frequency Domain Graph Convolutional Network for Traffic Forecasting}.
\newblock \bibinfo{journal}{\emph{arXiv preprint arXiv:2312.11933}} (\bibinfo{year}{2023}).
\newblock


\bibitem[Li et~al\mbox{.}(2018)]%
        {RN58}
\bibfield{author}{\bibinfo{person}{Yaguang Li}, \bibinfo{person}{Rose Yu}, \bibinfo{person}{Cyrus Shahabi}, {and} \bibinfo{person}{Yan Liu}.} \bibinfo{year}{2018}\natexlab{}.
\newblock \showarticletitle{Diffusion Convolutional Recurrent Neural Network: Data-Driven Traffic Forecasting}. In \bibinfo{booktitle}{\emph{International Conference on Learning Representations}}.
\newblock


\bibitem[Liang et~al\mbox{.}(2023a)]%
        {liang2023knowledge}
\bibfield{author}{\bibinfo{person}{Ke Liang}, \bibinfo{person}{Yue Liu}, \bibinfo{person}{Sihang Zhou}, \bibinfo{person}{Wenxuan Tu}, \bibinfo{person}{Yi Wen}, \bibinfo{person}{Xihong Yang}, \bibinfo{person}{Xiangjun Dong}, {and} \bibinfo{person}{Xinwang Liu}.} \bibinfo{year}{2023}\natexlab{a}.
\newblock \showarticletitle{Knowledge Graph Contrastive Learning Based on Relation-Symmetrical Structure}.
\newblock \bibinfo{journal}{\emph{IEEE Transactions on Knowledge and Data Engineering}} (\bibinfo{year}{2023}).
\newblock


\bibitem[Liang et~al\mbox{.}(2022)]%
        {liang2022reasoning}
\bibfield{author}{\bibinfo{person}{Ke Liang}, \bibinfo{person}{Lingyuan Meng}, \bibinfo{person}{Meng Liu}, \bibinfo{person}{Yue Liu}, \bibinfo{person}{Wenxuan Tu}, \bibinfo{person}{Siwei Wang}, \bibinfo{person}{Sihang Zhou}, \bibinfo{person}{Xinwang Liu}, {and} \bibinfo{person}{Fuchun Sun}.} \bibinfo{year}{2022}\natexlab{}.
\newblock \showarticletitle{Reasoning over different types of knowledge graphs: Static, temporal and multi-modal}.
\newblock \bibinfo{journal}{\emph{arXiv preprint arXiv:2212.05767}} (\bibinfo{year}{2022}), \bibinfo{pages}{7576--7584}.
\newblock


\bibitem[Liang et~al\mbox{.}(2023b)]%
        {liang2023abslearn}
\bibfield{author}{\bibinfo{person}{Ke Liang}, \bibinfo{person}{Jim Tan}, \bibinfo{person}{Dongrui Zeng}, \bibinfo{person}{Yongzhe Huang}, \bibinfo{person}{Xiaolei Huang}, {and} \bibinfo{person}{Gang Tan}.} \bibinfo{year}{2023}\natexlab{b}.
\newblock \showarticletitle{Abslearn: a gnn-based framework for aliasing and buffer-size information retrieval}.
\newblock \bibinfo{journal}{\emph{Pattern Analysis and Applications}} (\bibinfo{year}{2023}), \bibinfo{pages}{1--19}.
\newblock


\bibitem[Liu et~al\mbox{.}(2020)]%
        {RN275}
\bibfield{author}{\bibinfo{person}{Hui Liu}, \bibinfo{person}{Chengqing Yu}, \bibinfo{person}{Haiping Wu}, \bibinfo{person}{Zhu Duan}, {and} \bibinfo{person}{Guangxi Yan}.} \bibinfo{year}{2020}\natexlab{}.
\newblock \showarticletitle{A new hybrid ensemble deep reinforcement learning model for wind speed short term forecasting}.
\newblock \bibinfo{journal}{\emph{Energy}}  \bibinfo{volume}{202} (\bibinfo{year}{2020}), \bibinfo{pages}{117794}.
\newblock
\showISSN{0360-5442}


\bibitem[Liu et~al\mbox{.}(2021)]%
        {liu2021stochastic}
\bibfield{author}{\bibinfo{person}{Linfeng Liu}, \bibinfo{person}{Michael~C Hughes}, \bibinfo{person}{Soha Hassoun}, {and} \bibinfo{person}{Liping Liu}.} \bibinfo{year}{2021}\natexlab{}.
\newblock \showarticletitle{Stochastic iterative graph matching}. In \bibinfo{booktitle}{\emph{International Conference on Machine Learning}}. PMLR, \bibinfo{pages}{6815--6825}.
\newblock


\bibitem[Liu et~al\mbox{.}(2024)]%
        {liu2024rt}
\bibfield{author}{\bibinfo{person}{Yutian Liu}, \bibinfo{person}{Soora Rasouli}, \bibinfo{person}{Melvin Wong}, \bibinfo{person}{Tao Feng}, {and} \bibinfo{person}{Tianjin Huang}.} \bibinfo{year}{2024}\natexlab{}.
\newblock \showarticletitle{RT-GCN: Gaussian-based spatiotemporal graph convolutional network for robust traffic prediction}.
\newblock \bibinfo{journal}{\emph{Information Fusion}}  \bibinfo{volume}{102} (\bibinfo{year}{2024}), \bibinfo{pages}{102078}.
\newblock


\bibitem[Luo et~al\mbox{.}(2019)]%
        {luo2019e2gan}
\bibfield{author}{\bibinfo{person}{Yonghong Luo}, \bibinfo{person}{Ying Zhang}, \bibinfo{person}{Xiangrui Cai}, {and} \bibinfo{person}{Xiaojie Yuan}.} \bibinfo{year}{2019}\natexlab{}.
\newblock \showarticletitle{E2gan: End-to-end generative adversarial network for multivariate time series imputation}. In \bibinfo{booktitle}{\emph{Proceedings of the 28th international joint conference on artificial intelligence}}. AAAI Press Palo Alto, CA, USA, \bibinfo{pages}{3094--3100}.
\newblock


\bibitem[Pachal and Achar(2022)]%
        {pachal2022sequence}
\bibfield{author}{\bibinfo{person}{Soumen Pachal} {and} \bibinfo{person}{Avinash Achar}.} \bibinfo{year}{2022}\natexlab{}.
\newblock \showarticletitle{Sequence Prediction under Missing Data: An RNN Approach without Imputation}. In \bibinfo{booktitle}{\emph{Proceedings of the 31st ACM International Conference on Information \& Knowledge Management}}. \bibinfo{pages}{1605--1614}.
\newblock


\bibitem[Qian et~al\mbox{.}(2023)]%
        {qian2023adaptraj}
\bibfield{author}{\bibinfo{person}{Tangwen Qian}, \bibinfo{person}{Yile Chen}, \bibinfo{person}{Gao Cong}, \bibinfo{person}{Yongjun Xu}, {and} \bibinfo{person}{Fei Wang}.} \bibinfo{year}{2023}\natexlab{}.
\newblock \showarticletitle{AdapTraj: A Multi-Source Domain Generalization Framework for Multi-Agent Trajectory Prediction}.
\newblock \bibinfo{journal}{\emph{arXiv preprint arXiv:2312.14394}} (\bibinfo{year}{2023}).
\newblock


\bibitem[Ren et~al\mbox{.}(2023)]%
        {ren2023damr}
\bibfield{author}{\bibinfo{person}{Xiaobin Ren}, \bibinfo{person}{Kaiqi Zhao}, \bibinfo{person}{Patricia~J Riddle}, \bibinfo{person}{Katerina Taskova}, \bibinfo{person}{Qingyi Pan}, {and} \bibinfo{person}{Lianyan Li}.} \bibinfo{year}{2023}\natexlab{}.
\newblock \showarticletitle{DAMR: Dynamic Adjacency Matrix Representation Learning for Multivariate Time Series Imputation}.
\newblock \bibinfo{journal}{\emph{Proceedings of the ACM on Management of Data}} \bibinfo{volume}{1}, \bibinfo{number}{2} (\bibinfo{year}{2023}), \bibinfo{pages}{1--25}.
\newblock


\bibitem[Shan et~al\mbox{.}(2023)]%
        {shan2023nrtsi}
\bibfield{author}{\bibinfo{person}{Siyuan Shan}, \bibinfo{person}{Yang Li}, {and} \bibinfo{person}{Junier~B Oliva}.} \bibinfo{year}{2023}\natexlab{}.
\newblock \showarticletitle{Nrtsi: Non-recurrent time series imputation}. In \bibinfo{booktitle}{\emph{ICASSP 2023-2023 IEEE International Conference on Acoustics, Speech and Signal Processing (ICASSP)}}. IEEE, \bibinfo{pages}{1--5}.
\newblock


\bibitem[Shang et~al\mbox{.}(2021)]%
        {shang2021discrete}
\bibfield{author}{\bibinfo{person}{Chao Shang}, \bibinfo{person}{Jie Chen}, {and} \bibinfo{person}{Jinbo Bi}.} \bibinfo{year}{2021}\natexlab{}.
\newblock \showarticletitle{Discrete Graph Structure Learning for Forecasting Multiple Time Series}. In \bibinfo{booktitle}{\emph{International Conference on Learning Representations}}.
\newblock


\bibitem[Shang et~al\mbox{.}(2022)]%
        {shang2022new}
\bibfield{author}{\bibinfo{person}{Pan Shang}, \bibinfo{person}{Xinwei Liu}, \bibinfo{person}{Chengqing Yu}, \bibinfo{person}{Guangxi Yan}, \bibinfo{person}{Qingqing Xiang}, {and} \bibinfo{person}{Xiwei Mi}.} \bibinfo{year}{2022}\natexlab{}.
\newblock \showarticletitle{A new ensemble deep graph reinforcement learning network for spatio-temporal traffic volume forecasting in a freeway network}.
\newblock \bibinfo{journal}{\emph{Digital Signal Processing}}  \bibinfo{volume}{123} (\bibinfo{year}{2022}), \bibinfo{pages}{103419}.
\newblock


\bibitem[Shao et~al\mbox{.}(2023)]%
        {shao2023exploring}
\bibfield{author}{\bibinfo{person}{Zezhi Shao}, \bibinfo{person}{Fei Wang}, \bibinfo{person}{Yongjun Xu}, \bibinfo{person}{Wei Wei}, \bibinfo{person}{Chengqing Yu}, \bibinfo{person}{Zhao Zhang}, \bibinfo{person}{Di Yao}, \bibinfo{person}{Guangyin Jin}, \bibinfo{person}{Xin Cao}, \bibinfo{person}{Gao Cong}, {et~al\mbox{.}}} \bibinfo{year}{2023}\natexlab{}.
\newblock \showarticletitle{Exploring Progress in Multivariate Time Series Forecasting: Comprehensive Benchmarking and Heterogeneity Analysis}.
\newblock \bibinfo{journal}{\emph{arXiv preprint arXiv:2310.06119}} (\bibinfo{year}{2023}).
\newblock


\bibitem[Shao et~al\mbox{.}(2022b)]%
        {RN859}
\bibfield{author}{\bibinfo{person}{Zezhi Shao}, \bibinfo{person}{Zhao Zhang}, \bibinfo{person}{Fei Wang}, \bibinfo{person}{Wei Wei}, {and} \bibinfo{person}{Yongjun Xu}.} \bibinfo{year}{2022}\natexlab{b}.
\newblock \showarticletitle{Spatial-Temporal Identity: A Simple yet Effective Baseline for Multivariate Time Series Forecasting}. In \bibinfo{booktitle}{\emph{Proceedings of the 31st ACM International Conference on Information and Knowledge Management}}. \bibinfo{pages}{4454--4458}.
\newblock


\bibitem[Shao et~al\mbox{.}(2022a)]%
        {shao2022pre}
\bibfield{author}{\bibinfo{person}{Zezhi Shao}, \bibinfo{person}{Zhao Zhang}, \bibinfo{person}{Fei Wang}, {and} \bibinfo{person}{Yongjun Xu}.} \bibinfo{year}{2022}\natexlab{a}.
\newblock \showarticletitle{Pre-training enhanced spatial-temporal graph neural network for multivariate time series forecasting}. In \bibinfo{booktitle}{\emph{Proceedings of the 28th ACM SIGKDD Conference on Knowledge Discovery and Data Mining}}. \bibinfo{pages}{1567--1577}.
\newblock


\bibitem[Shao et~al\mbox{.}(2022c)]%
        {shao2022decoupled}
\bibfield{author}{\bibinfo{person}{Zezhi Shao}, \bibinfo{person}{Zhao Zhang}, \bibinfo{person}{Wei Wei}, \bibinfo{person}{Fei Wang}, \bibinfo{person}{Yongjun Xu}, \bibinfo{person}{Xin Cao}, {and} \bibinfo{person}{Christian~S Jensen}.} \bibinfo{year}{2022}\natexlab{c}.
\newblock \showarticletitle{Decoupled dynamic spatial-temporal graph neural network for traffic forecasting}.
\newblock \bibinfo{journal}{\emph{Proceedings of the VLDB Endowment}} \bibinfo{volume}{15}, \bibinfo{number}{11} (\bibinfo{year}{2022}), \bibinfo{pages}{2733--2746}.
\newblock


\bibitem[Su et~al\mbox{.}(2023)]%
        {su2023novel}
\bibfield{author}{\bibinfo{person}{Mengshuai Su}, \bibinfo{person}{Hui Liu}, \bibinfo{person}{Chengqing Yu}, {and} \bibinfo{person}{Zhu Duan}.} \bibinfo{year}{2023}\natexlab{}.
\newblock \showarticletitle{A novel AQI forecasting method based on fusing temporal correlation forecasting with spatial correlation forecasting}.
\newblock \bibinfo{journal}{\emph{Atmospheric Pollution Research}} \bibinfo{volume}{14}, \bibinfo{number}{4} (\bibinfo{year}{2023}), \bibinfo{pages}{101717}.
\newblock


\bibitem[Sun et~al\mbox{.}(2022)]%
        {sun2022human}
\bibfield{author}{\bibinfo{person}{Tao Sun}, \bibinfo{person}{Fei Wang}, \bibinfo{person}{Zhao Zhang}, \bibinfo{person}{Lin Wu}, {and} \bibinfo{person}{Yongjun Xu}.} \bibinfo{year}{2022}\natexlab{}.
\newblock \showarticletitle{Human mobility identification by deep behavior relevant location representation}. In \bibinfo{booktitle}{\emph{International Conference on Database Systems for Advanced Applications}}. Springer, \bibinfo{pages}{439--454}.
\newblock


\bibitem[Tan et~al\mbox{.}(2022)]%
        {tan2022new}
\bibfield{author}{\bibinfo{person}{Jing Tan}, \bibinfo{person}{Hui Liu}, \bibinfo{person}{Yanfei Li}, \bibinfo{person}{Shi Yin}, {and} \bibinfo{person}{Chengqing Yu}.} \bibinfo{year}{2022}\natexlab{}.
\newblock \showarticletitle{A new ensemble spatio-temporal PM2. 5 prediction method based on graph attention recursive networks and reinforcement learning}.
\newblock \bibinfo{journal}{\emph{Chaos, Solitons \& Fractals}}  \bibinfo{volume}{162} (\bibinfo{year}{2022}), \bibinfo{pages}{112405}.
\newblock


\bibitem[Tang et~al\mbox{.}(2023)]%
        {tang2023recurrent}
\bibfield{author}{\bibinfo{person}{Peiwang Tang}, \bibinfo{person}{Qinghua Zhang}, {and} \bibinfo{person}{Xianchao Zhang}.} \bibinfo{year}{2023}\natexlab{}.
\newblock \showarticletitle{A Recurrent Neural Network based Generative Adversarial Network for Long Multivariate Time Series Forecasting}. In \bibinfo{booktitle}{\emph{Proceedings of the 2023 ACM International Conference on Multimedia Retrieval}}. \bibinfo{pages}{181--189}.
\newblock


\bibitem[Tang et~al\mbox{.}(2020)]%
        {tang2020joint}
\bibfield{author}{\bibinfo{person}{Xianfeng Tang}, \bibinfo{person}{Huaxiu Yao}, \bibinfo{person}{Yiwei Sun}, \bibinfo{person}{Charu Aggarwal}, \bibinfo{person}{Prasenjit Mitra}, {and} \bibinfo{person}{Suhang Wang}.} \bibinfo{year}{2020}\natexlab{}.
\newblock \showarticletitle{Joint modeling of local and global temporal dynamics for multivariate time series forecasting with missing values}. In \bibinfo{booktitle}{\emph{Proceedings of the AAAI Conference on Artificial Intelligence}}, Vol.~\bibinfo{volume}{34}. \bibinfo{pages}{5956--5963}.
\newblock


\bibitem[Wang et~al\mbox{.}(2023a)]%
        {wang2023ai}
\bibfield{author}{\bibinfo{person}{Fei Wang}, \bibinfo{person}{Di Yao}, \bibinfo{person}{Yong Li}, \bibinfo{person}{Tao Sun}, {and} \bibinfo{person}{Zhao Zhang}.} \bibinfo{year}{2023}\natexlab{a}.
\newblock \showarticletitle{AI-enhanced spatial-temporal data-mining technology: New chance for next-generation urban computing}.
\newblock \bibinfo{journal}{\emph{The Innovation}} \bibinfo{volume}{4}, \bibinfo{number}{2} (\bibinfo{year}{2023}).
\newblock


\bibitem[Wang et~al\mbox{.}(2019)]%
        {wang2019fingerprint}
\bibfield{author}{\bibinfo{person}{Pu Wang}, \bibinfo{person}{Zhihong Feng}, \bibinfo{person}{Yan Tang}, {and} \bibinfo{person}{Yuzhi Zhang}.} \bibinfo{year}{2019}\natexlab{}.
\newblock \showarticletitle{A fingerprint database reconstruction method based on ordinary kriging algorithm for indoor localization}. In \bibinfo{booktitle}{\emph{2019 International Conference on Intelligent Transportation, Big Data \& Smart City (ICITBS)}}. IEEE, \bibinfo{pages}{224--227}.
\newblock


\bibitem[Wang et~al\mbox{.}(2022)]%
        {wang2022multi}
\bibfield{author}{\bibinfo{person}{Peixiao Wang}, \bibinfo{person}{Tong Zhang}, \bibinfo{person}{Yueming Zheng}, {and} \bibinfo{person}{Tao Hu}.} \bibinfo{year}{2022}\natexlab{}.
\newblock \showarticletitle{A multi-view bidirectional spatiotemporal graph network for urban traffic flow imputation}.
\newblock \bibinfo{journal}{\emph{International Journal of Geographical Information Science}} \bibinfo{volume}{36}, \bibinfo{number}{6} (\bibinfo{year}{2022}), \bibinfo{pages}{1231--1257}.
\newblock


\bibitem[Wang et~al\mbox{.}(2023b)]%
        {wang2023learning}
\bibfield{author}{\bibinfo{person}{Zhiyuan Wang}, \bibinfo{person}{Fan Zhou}, \bibinfo{person}{Goce Trajcevski}, \bibinfo{person}{Kunpeng Zhang}, {and} \bibinfo{person}{Ting Zhong}.} \bibinfo{year}{2023}\natexlab{b}.
\newblock \showarticletitle{Learning Dynamic Temporal Relations with Continuous Graph for Multivariate Time Series Forecasting (Student Abstract)}. In \bibinfo{booktitle}{\emph{Proceedings of the AAAI Conference on Artificial Intelligence}}, Vol.~\bibinfo{volume}{37}. \bibinfo{pages}{16358--16359}.
\newblock


\bibitem[Wei et~al\mbox{.}(2023)]%
        {wei2023lstm}
\bibfield{author}{\bibinfo{person}{Yuanyuan Wei}, \bibinfo{person}{Julian Jang-Jaccard}, \bibinfo{person}{Wen Xu}, \bibinfo{person}{Fariza Sabrina}, \bibinfo{person}{Seyit Camtepe}, {and} \bibinfo{person}{Mikael Boulic}.} \bibinfo{year}{2023}\natexlab{}.
\newblock \showarticletitle{LSTM-autoencoder-based anomaly detection for indoor air quality time-series data}.
\newblock \bibinfo{journal}{\emph{IEEE Sensors Journal}} \bibinfo{volume}{23}, \bibinfo{number}{4} (\bibinfo{year}{2023}), \bibinfo{pages}{3787--3800}.
\newblock


\bibitem[Wu et~al\mbox{.}(2023)]%
        {RN18}
\bibfield{author}{\bibinfo{person}{Haixu Wu}, \bibinfo{person}{Tengge Hu}, \bibinfo{person}{Yong Liu}, \bibinfo{person}{Hang Zhou}, \bibinfo{person}{Jianmin Wang}, {and} \bibinfo{person}{Mingsheng Long}.} \bibinfo{year}{2023}\natexlab{}.
\newblock \showarticletitle{TimesNet: Temporal 2D-Variation Modeling for General Time Series Analysis}. In \bibinfo{booktitle}{\emph{The Eleventh International Conference on Learning Representations}}.
\newblock


\bibitem[Wu et~al\mbox{.}(2021a)]%
        {wu2021autoformer}
\bibfield{author}{\bibinfo{person}{Haixu Wu}, \bibinfo{person}{Jiehui Xu}, \bibinfo{person}{Jianmin Wang}, {and} \bibinfo{person}{Mingsheng Long}.} \bibinfo{year}{2021}\natexlab{a}.
\newblock \showarticletitle{Autoformer: Decomposition transformers with auto-correlation for long-term series forecasting}.
\newblock \bibinfo{journal}{\emph{Advances in Neural Information Processing Systems}}  \bibinfo{volume}{34} (\bibinfo{year}{2021}), \bibinfo{pages}{22419--22430}.
\newblock


\bibitem[Wu et~al\mbox{.}(2021b)]%
        {wu2021inductive}
\bibfield{author}{\bibinfo{person}{Yuankai Wu}, \bibinfo{person}{Dingyi Zhuang}, \bibinfo{person}{Aurelie Labbe}, {and} \bibinfo{person}{Lijun Sun}.} \bibinfo{year}{2021}\natexlab{b}.
\newblock \showarticletitle{Inductive graph neural networks for spatiotemporal kriging}. In \bibinfo{booktitle}{\emph{Proceedings of the AAAI Conference on Artificial Intelligence}}, Vol.~\bibinfo{volume}{35}. \bibinfo{pages}{4478--4485}.
\newblock


\bibitem[Wu et~al\mbox{.}(2020)]%
        {RN55}
\bibfield{author}{\bibinfo{person}{Zonghan Wu}, \bibinfo{person}{Shirui Pan}, \bibinfo{person}{Guodong Long}, \bibinfo{person}{Jing Jiang}, \bibinfo{person}{Xiaojun Chang}, {and} \bibinfo{person}{Chengqi Zhang}.} \bibinfo{year}{2020}\natexlab{}.
\newblock \showarticletitle{Connecting the dots: Multivariate time series forecasting with graph neural networks}. In \bibinfo{booktitle}{\emph{Proceedings of the 26th ACM SIGKDD international conference on knowledge discovery \& data mining}}. \bibinfo{pages}{753--763}.
\newblock


\bibitem[Wu et~al\mbox{.}(2019)]%
        {RN56}
\bibfield{author}{\bibinfo{person}{Zonghan Wu}, \bibinfo{person}{Shirui Pan}, \bibinfo{person}{Guodong Long}, \bibinfo{person}{Jing Jiang}, {and} \bibinfo{person}{Chengqi Zhang}.} \bibinfo{year}{2019}\natexlab{}.
\newblock \showarticletitle{Graph wavenet for deep spatial-temporal graph modeling}. In \bibinfo{booktitle}{\emph{Proceedings of the 28th International Joint Conference on Artificial Intelligence}}. \bibinfo{pages}{1907--1913}.
\newblock


\bibitem[Xu et~al\mbox{.}(2023a)]%
        {xu2023uncovering}
\bibfield{author}{\bibinfo{person}{Yi Xu}, \bibinfo{person}{Armin Bazarjani}, \bibinfo{person}{Hyung-gun Chi}, \bibinfo{person}{Chiho Choi}, {and} \bibinfo{person}{Yun Fu}.} \bibinfo{year}{2023}\natexlab{a}.
\newblock \showarticletitle{Uncovering the Missing Pattern: Unified Framework Towards Trajectory Imputation and Prediction}. In \bibinfo{booktitle}{\emph{Proceedings of the IEEE/CVF Conference on Computer Vision and Pattern Recognition}}. \bibinfo{pages}{9632--9643}.
\newblock


\bibitem[Xu et~al\mbox{.}(2021)]%
        {xu2021artificial}
\bibfield{author}{\bibinfo{person}{Yongjun Xu}, \bibinfo{person}{Xin Liu}, \bibinfo{person}{Xin Cao}, \bibinfo{person}{Changping Huang}, \bibinfo{person}{Enke Liu}, \bibinfo{person}{Sen Qian}, \bibinfo{person}{Xingchen Liu}, \bibinfo{person}{Yanjun Wu}, \bibinfo{person}{Fengliang Dong}, \bibinfo{person}{Cheng-Wei Qiu}, {et~al\mbox{.}}} \bibinfo{year}{2021}\natexlab{}.
\newblock \showarticletitle{Artificial intelligence: A powerful paradigm for scientific research}.
\newblock \bibinfo{journal}{\emph{The Innovation}} \bibinfo{volume}{2}, \bibinfo{number}{4} (\bibinfo{year}{2021}), \bibinfo{pages}{100179}.
\newblock


\bibitem[Xu et~al\mbox{.}(2023b)]%
        {xu2023artificial}
\bibfield{author}{\bibinfo{person}{Yongjun Xu}, \bibinfo{person}{Fei Wang}, \bibinfo{person}{Zhulin An}, \bibinfo{person}{Qi Wang}, {and} \bibinfo{person}{Zhao Zhang}.} \bibinfo{year}{2023}\natexlab{b}.
\newblock \showarticletitle{Artificial intelligence for science—bridging data to wisdom}.
\newblock \bibinfo{journal}{\emph{The Innovation}} \bibinfo{volume}{4}, \bibinfo{number}{6} (\bibinfo{year}{2023}).
\newblock


\bibitem[Ye et~al\mbox{.}(2021)]%
        {ye2021spatial}
\bibfield{author}{\bibinfo{person}{Yongchao Ye}, \bibinfo{person}{Shiyao Zhang}, {and} \bibinfo{person}{James~JQ Yu}.} \bibinfo{year}{2021}\natexlab{}.
\newblock \showarticletitle{Spatial-temporal traffic data imputation via graph attention convolutional network}. In \bibinfo{booktitle}{\emph{International Conference on Artificial Neural Networks}}. Springer, \bibinfo{pages}{241--252}.
\newblock


\bibitem[Yi et~al\mbox{.}(2023)]%
        {yi2023fouriergnn}
\bibfield{author}{\bibinfo{person}{Kun Yi}, \bibinfo{person}{Qi Zhang}, \bibinfo{person}{Wei Fan}, \bibinfo{person}{Hui He}, \bibinfo{person}{Liang Hu}, \bibinfo{person}{Pengyang Wang}, \bibinfo{person}{Ning An}, \bibinfo{person}{Longbing Cao}, {and} \bibinfo{person}{Zhendong Niu}.} \bibinfo{year}{2023}\natexlab{}.
\newblock \showarticletitle{FourierGNN: Rethinking Multivariate Time Series Forecasting from a Pure Graph Perspective}. In \bibinfo{booktitle}{\emph{Thirty-seventh Conference on Neural Information Processing Systems}}.
\newblock


\bibitem[Yick et~al\mbox{.}(2008)]%
        {yick2008wireless}
\bibfield{author}{\bibinfo{person}{Jennifer Yick}, \bibinfo{person}{Biswanath Mukherjee}, {and} \bibinfo{person}{Dipak Ghosal}.} \bibinfo{year}{2008}\natexlab{}.
\newblock \showarticletitle{Wireless sensor network survey}.
\newblock \bibinfo{journal}{\emph{Computer networks}} \bibinfo{volume}{52}, \bibinfo{number}{12} (\bibinfo{year}{2008}), \bibinfo{pages}{2292--2330}.
\newblock


\bibitem[Yin et~al\mbox{.}(2023)]%
        {yin2023mtmgnn}
\bibfield{author}{\bibinfo{person}{Du Yin}, \bibinfo{person}{Renhe Jiang}, \bibinfo{person}{Jiewen Deng}, \bibinfo{person}{Yongkang Li}, \bibinfo{person}{Yi Xie}, \bibinfo{person}{Zhongyi Wang}, \bibinfo{person}{Yifan Zhou}, \bibinfo{person}{Xuan Song}, {and} \bibinfo{person}{Jedi~S Shang}.} \bibinfo{year}{2023}\natexlab{}.
\newblock \showarticletitle{MTMGNN: Multi-time multi-graph neural network for metro passenger flow prediction}.
\newblock \bibinfo{journal}{\emph{GeoInformatica}} \bibinfo{volume}{27}, \bibinfo{number}{1} (\bibinfo{year}{2023}), \bibinfo{pages}{77--105}.
\newblock


\bibitem[Yoon et~al\mbox{.}(2018)]%
        {yoon2018gain}
\bibfield{author}{\bibinfo{person}{Jinsung Yoon}, \bibinfo{person}{James Jordon}, {and} \bibinfo{person}{Mihaela Schaar}.} \bibinfo{year}{2018}\natexlab{}.
\newblock \showarticletitle{Gain: Missing data imputation using generative adversarial nets}. In \bibinfo{booktitle}{\emph{International conference on machine learning}}. PMLR, \bibinfo{pages}{5689--5698}.
\newblock


\bibitem[Yu et~al\mbox{.}(2023)]%
        {yu2023dsformer}
\bibfield{author}{\bibinfo{person}{Chengqing Yu}, \bibinfo{person}{Fei Wang}, \bibinfo{person}{Zezhi Shao}, \bibinfo{person}{Tao Sun}, \bibinfo{person}{Lin Wu}, {and} \bibinfo{person}{Yongjun Xu}.} \bibinfo{year}{2023}\natexlab{}.
\newblock \showarticletitle{Dsformer: A double sampling transformer for multivariate time series long-term prediction}. In \bibinfo{booktitle}{\emph{Proceedings of the 32nd ACM International Conference on Information and Knowledge Management}}. \bibinfo{pages}{3062--3072}.
\newblock


\bibitem[Yu et~al\mbox{.}(2024)]%
        {yu2023MRIformer}
\bibfield{author}{\bibinfo{person}{Chengqing Yu}, \bibinfo{person}{Guangxi Yan}, \bibinfo{person}{Chengming Yu}, \bibinfo{person}{Xinwei Liu}, {and} \bibinfo{person}{Xiwei Mi}.} \bibinfo{year}{2024}\natexlab{}.
\newblock \showarticletitle{MRIformer: A multi-resolution interactive transformer for wind speed multi-step prediction}.
\newblock \bibinfo{journal}{\emph{Information Sciences}}  \bibinfo{volume}{661} (\bibinfo{year}{2024}), \bibinfo{pages}{120150}.
\newblock


\bibitem[Zhang et~al\mbox{.}(2023)]%
        {zhang2023trid}
\bibfield{author}{\bibinfo{person}{Kai Zhang}, \bibinfo{person}{Chao Li}, {and} \bibinfo{person}{Qinmin Yang}.} \bibinfo{year}{2023}\natexlab{}.
\newblock \showarticletitle{TriD-MAE: A Generic Pre-trained Model for Multivariate Time Series with Missing Values}. In \bibinfo{booktitle}{\emph{Proceedings of the 32nd ACM International Conference on Information and Knowledge Management}}. \bibinfo{pages}{3164--3173}.
\newblock


\bibitem[Zhao et~al\mbox{.}(2019)]%
        {zhao2019t}
\bibfield{author}{\bibinfo{person}{Ling Zhao}, \bibinfo{person}{Yujiao Song}, \bibinfo{person}{Chao Zhang}, \bibinfo{person}{Yu Liu}, \bibinfo{person}{Pu Wang}, \bibinfo{person}{Tao Lin}, \bibinfo{person}{Min Deng}, {and} \bibinfo{person}{Haifeng Li}.} \bibinfo{year}{2019}\natexlab{}.
\newblock \showarticletitle{T-gcn: A temporal graph convolutional network for traffic prediction}.
\newblock \bibinfo{journal}{\emph{IEEE transactions on intelligent transportation systems}} \bibinfo{volume}{21}, \bibinfo{number}{9} (\bibinfo{year}{2019}), \bibinfo{pages}{3848--3858}.
\newblock


\bibitem[Zheng et~al\mbox{.}(2020)]%
        {zheng2020gman}
\bibfield{author}{\bibinfo{person}{Chuanpan Zheng}, \bibinfo{person}{Xiaoliang Fan}, \bibinfo{person}{Cheng Wang}, {and} \bibinfo{person}{Jianzhong Qi}.} \bibinfo{year}{2020}\natexlab{}.
\newblock \showarticletitle{Gman: A graph multi-attention network for traffic prediction}. In \bibinfo{booktitle}{\emph{Proceedings of the AAAI conference on artificial intelligence}}, Vol.~\bibinfo{volume}{34}. \bibinfo{pages}{1234--1241}.
\newblock


\bibitem[Zhou et~al\mbox{.}(2023b)]%
        {zhou2023sloth}
\bibfield{author}{\bibinfo{person}{Fan Zhou}, \bibinfo{person}{Chen Pan}, \bibinfo{person}{Lintao Ma}, \bibinfo{person}{Yu Liu}, \bibinfo{person}{Shiyu Wang}, \bibinfo{person}{James Zhang}, \bibinfo{person}{Xinxin Zhu}, \bibinfo{person}{Xuanwei Hu}, \bibinfo{person}{Yunhua Hu}, \bibinfo{person}{Yangfei Zheng}, {et~al\mbox{.}}} \bibinfo{year}{2023}\natexlab{b}.
\newblock \showarticletitle{SLOTH: Structured Learning and Task-Based Optimization for Time Series Forecasting on Hierarchies}. In \bibinfo{booktitle}{\emph{Proceedings of the AAAI Conference on Artificial Intelligence}}, Vol.~\bibinfo{volume}{37}. \bibinfo{pages}{11417--11425}.
\newblock


\bibitem[Zhou et~al\mbox{.}(2023a)]%
        {zhou2023one}
\bibfield{author}{\bibinfo{person}{Tian Zhou}, \bibinfo{person}{Peisong Niu}, \bibinfo{person}{Xue Wang}, \bibinfo{person}{Liang Sun}, {and} \bibinfo{person}{Rong Jin}.} \bibinfo{year}{2023}\natexlab{a}.
\newblock \showarticletitle{One Fits All: Universal Time Series Analysis by Pretrained LM and Specially Designed Adaptors}.
\newblock \bibinfo{journal}{\emph{arXiv preprint arXiv:2311.14782}} (\bibinfo{year}{2023}).
\newblock


\end{thebibliography}

\appendix

\section{Experimental details}

\subsection{Datasets}
\autoref{tab2} shows the statistics of these datasets. A brief overview of all datasets is shown as follows:
\begin{itemize}
    \item \textbf{METR-LA}: It is a traffic speed dataset collected by loop-detectors located on the LA County road network, which contains data collected by 207 sensors from Mar 1st, 2012 to Jun 30th, 2012. Each time series is sampled at a 5-minute interval, totaling 34272 time slices.
    \item \textbf{PEMS-BAY}: It is a traffic speed dataset collected by California Transportation Agencies (CalTrans) Performance Measurement System (PeMS), which contains data collected by 325 sensors from Jan 1st, 2017 to May 31th, 2017. Each time series is sampled at a 5-minute interval, totaling 52116 time slices.
    \item \textbf{PEMS04}: It is a traffic flow dataset collected by CalTrans PeMS, which contains data collected by 307 sensors from January 1st, 2018 to February 28th, 2018. Each time series is sampled at a 5-minute interval, totaling 16992 time slices.
    \item \textbf{PEMS08}: It is a traffic flow dataset collected by CalTrans PeMS, which contains data collected by 170 sensors from July 1st, 2018 to Aug 31th, 2018. Each time series is sampled at a 5-minute interval, totaling 17833 time slices.
    \item \textbf{China AQI}: It is an air quality dataset collected by China Environmental Monitoring Station, which contains data collected by 350 cities in China from January 2015 to December 2022. Each time series is sampled at a 1 hour interval, totaling 59710 time slices.
\end{itemize}

\begin{table}
\centering
  \caption{The statistics of the five datasets.}
  \label{tab2}
  \begin{tabular}{cccc}
\toprule
    Datasets&Variates &Timesteps&Granularity\\
\midrule
MET-LA&207&34272&5minutes\\
PEMS-BAY& 325&52116&5minutes\\
PEMS04& 307&16992&5minutes\\
PEMS08&170&17856&5minutes\\
China AQI&350&59710&1hour\\
\toprule
\end{tabular}
\end{table}

\subsection{Baselines}
 All baselines are introduced as follows:
\begin{itemize}
    \item \textbf{STID}: It uses spatial-temporal identity embedding to improve the ability of MLP to mine multivariate time series.
    \item \textbf{DSformer}: It uses double sampling block and temproal variable attention block to mine spatiotemporal correlation and improve prediction performance.
    \item \textbf{MegaCRN}: This method uses memory back to improve the ability of AGCRN to model spatial correlation.
    \item \textbf{DCRNN}+\textbf{GPT4TS} : It first uses GPT4TS to realize the imputation of missing variables, and then uses DCRNN to model the processed data.
    \item \textbf{DFDGCN} + \textbf{TimesNet}: It first uses TimesNet to realize the imputation of missing variables, and then uses DFDGCN to model the processed data.
    \item \textbf{MTGNN}+\textbf{GRIN}: It first uses GRIN to realize the imputation of missing variables, and then uses MTGNN to model the processed data.
    \item \textbf{FourierGNN}+\textbf{GATGPT} :  It first uses GATGPT to realize the imputation of missing variables, and then uses FourierGNN to model the processed data.
    \item \textbf{LGnet}: It uses the memory component to effectively improve the performance of long and short term memory networks.
    \item \textbf{GC-VRNN} : It combines the Multi-Space Graph Neural Network with Conditional Variational Recurrent Neural to realize time series forecasting with missing values.
    \item \textbf{TriD-MAE}: It uses MAE to optimize the ability of the TCN model to realize multivariate time series forecasting with missing values.
    \item \textbf{BiTGraph}: It proposes a Biased Temporal Convolution Graph Network that jointly captures the temporal dependencies and spatial structure.
    
\end{itemize}

\section{Efficiency}

\begin{figure}[h]
\center
\includegraphics[width=\linewidth]{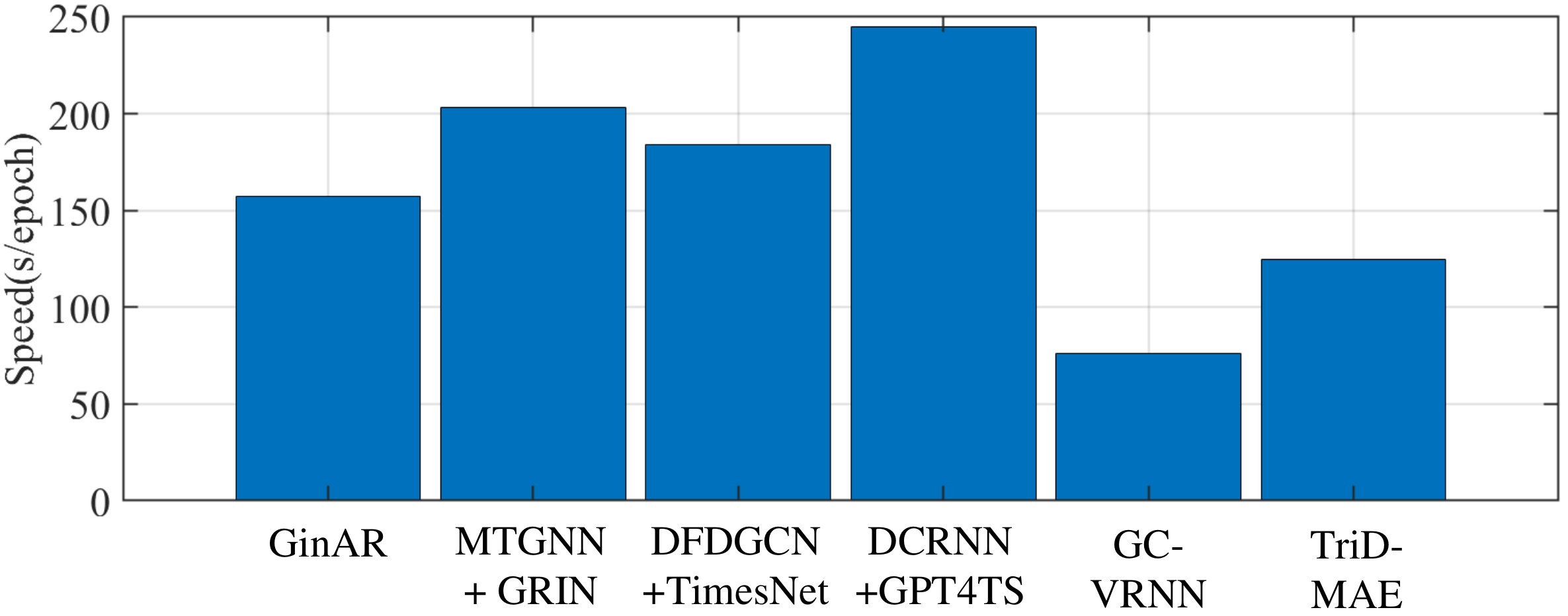}
\caption{Training time for each epoch of different models. }
\label{fig7}
\end{figure}

In this section, we compare the efficiency of GinAR with that of several baselines (GC-VRNN, TriD-MAE, MTGNN + GRIN, DFDGCN + TimesNet and DCRNN + GPT4TS) on the PEMS08 dataset. To ensure the fairness of the experiment, we compare the mean training time of each epoch of each model. The experimental equipment is the Intel(R) Xeon(R) Gold 5217 CPU @ 3.00GHz, 128G RAM computing server with RTX 3090 graphics card. The batch size is set to 16. Based on \autoref{fig7}, it can be found that the training time of GinAR is not large. Compared with several two-stage models, the GinAR does not require the imputation stage, which reduces the overall training time. Besides, although the training time of GinAR is greater than that of the one-stage models, it solves the problem of variable missing, which can improve its forecasting performance.

\section{Notations}
Some of the commonly used notations are presented in \autoref{tab1}.
\begin{table}[h]
\small
\centering
  \caption{Frequently used notation.}
  \label{tab1}
  \begin{tabular}{ccc}
\toprule
    Notation&size&Definitions\\
\midrule
$H$& Constant&The length of historical observation\\
$L$& Constant&The length of future forecasting results\\
$B$&Constant&Batch size\\
$N$&Constant&Number of variables\\
$M$&Constant&Number of missing variables\\
$C$&Constant&Embedding size\\
$d$&Constant&Variable embedding size\\
$n$&Constant&Number of GinAR layers\\
$X$&$N*H*C$&Input features\\
$X_M$&$N*H*C$&Input features with $M$ missing variables\\
$Y$&$N*L$&Forecasting results\\
$A_{pre}$&$N*N$&Predefined graph\\
$A_{adap}$&$N*N$&Adaptive graph\\
$E_{a}$&$N*d$&Variable embedding of adaptive graph\\
$F_{LN}$&Functions&Layer normalization\\
$FC$&Functions&Fully connected layer\\
\text{ReLU}&Functions&Activation function ReLU\\
\text{ELU}&Functions&Activation function ELU\\
\text{GeLU}&Functions&Activation function GeLU\\
\text{LeakyReLU}&Functions&Activation function LeakyReLU\\
\text{softmax}&Functions&Activation function softmax\\
\toprule
\end{tabular}
\end{table}

\end{document}